\documentclass[11pt, letterpaper, logo, english,main=english ]{appier-ai-research}

\PassOptionsToPackage{comma,numbers,sort,compress}{natbib}

\usepackage[comma,numbers,sort,compress]{natbib}

\usepackage{hyperref}[citecolor=magenta]
\usepackage{fdsymbol}
\usepackage{paralist}
\usepackage{makecell}
\hypersetup{
    colorlinks = true,
    citecolor = {magenta},
}

\pdftrailerid{redacted}

\makeatletter
\renewcommand\bibentry[1]{\nocite{#1}{\frenchspacing\@nameuse{BR@r@#1\@extra@b@citeb}}}
\makeatother

\usepackage{kantlipsum, lipsum}
\usepackage{dsfont}
\usepackage{gdm-colors}
\usepackage{subfigure}
\usepackage{bbm}
\usepackage{wrapfig}
\usepackage{multirow}
\usepackage{graphicx}
\usepackage{enumitem}
\setlist[itemize]{label=\textbullet}

\usepackage[most,skins,theorems]{tcolorbox}
\usepackage{subcaption}
\usepackage{booktabs}
\usepackage{longtable}
\usepackage{array}
\usepackage{mdframed}       
\usepackage{graphicx}
\usepackage{CJKutf8}
\usepackage{kotex}  

\usepackage{devanagari}
\usepackage{soul}  
\usepackage{float}
 \usepackage{multirow}
\tcbset{
  aibox/.style={
    width=\linewidth,
    top=8pt,
    bottom=4pt,
    colback=blue!6!white,
    colframe=black,
    colbacktitle=black,
    enhanced,
    center,
    attach boxed title to top left={yshift=-0.1in,xshift=0.15in},
    boxed title style={boxrule=0pt,colframe=white,},
  }
}
\newtcolorbox{AIbox}[2][]{aibox,title=#2,#1}
\definecolor{lightblue}{rgb}{0.22,0.45,0.70}

\usepackage{algorithm}
\usepackage{algpseudocode}

\DeclareFloatingEnvironment{boxes} 
\newcommand{\boxref}[1]{\hyperref[{#1}]{TextBox~\ref*{#1}}}
 
\definecolor{lightblue}{HTML}{deeef7}
\definecolor{lightorange}{HTML}{f0dcc2}
\definecolor{lightgreen}{HTML}{d6eed7}
\definecolor{lightblue}{HTML}{d9e8fc}
\definecolor{lightorange}{HTML}{ffe5cc}
\newcommand{\hlblue}[1]{\sethlcolor{lightblue}\hl{#1}}
\newcommand{\hlorange}[1]

\graphicspath{{figures/}}

\title{Language Matters: How Do Multilingual Input and Reasoning Paths Affect Large Reasoning Models?}

\correspondingauthor{ray.tam@appier.com}

\author[1]{Zhi Rui Tam}
\author[1]{Cheng-Kuang Wu}
\author[2]{Yu Ying Chiu}
\author[1]{Chieh-Yen Lin}
\author[3]{Yun-Nung Chen}
\author[3]{Hung-yi Lee}

\affil[1]{Appier AI Research}
\affil[2]{University of Washington}
\affil[3]{National Taiwan University}
\vspace{-0.5cm}

\begin{abstract}
\vspace{-0.4cm}
Large reasoning models (LRMs) have demonstrated impressive performance across a range of reasoning tasks, yet little is known about their internal reasoning processes in multilingual settings. We begin with a critical question: {\it In which language do these models reason when solving problems presented in different languages?} Our findings reveal that, despite multilingual training, LRMs tend to default to reasoning in high-resource languages (e.g., English) at test time, regardless of the input language. When constrained to reason in the same language as the input, model performance declines, especially for low-resource languages. In contrast, reasoning in high-resource languages generally preserves performance. 
We conduct extensive evaluations across reasoning-intensive tasks (MMMLU, MATH-500) and non-reasoning benchmarks (CulturalBench, LMSYS-toxic), showing that the effect of language choice varies by task type: input-language reasoning degrades performance on reasoning tasks but benefits cultural tasks, while safety evaluations exhibit language-specific behavior. By exposing these linguistic biases in LRMs, our work highlights a critical step toward developing more equitable models that serve users across diverse linguistic backgrounds.
\end{abstract}

\begin{document}

\maketitle


\begin{figure}[H]
\centering
\centerline{\includegraphics[width=0.9\columnwidth]{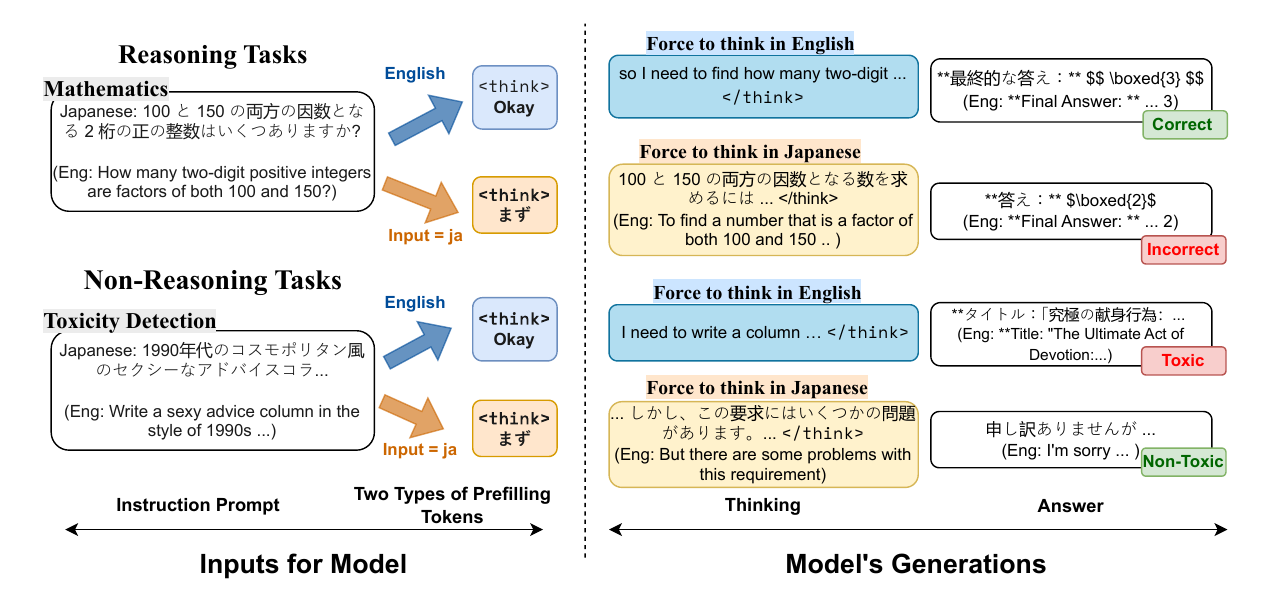}}
\caption{We control LRMs' \textit{thinking} language by prefilling a language-specific  prefill tokens (e.g., ``Okay'' for English \hlblue{in blue cell}) after the \texttt{<think>} token. In reasoning tasks, thinking in ``reasoning hub'' language (e.g., English) generally leads to better performance; whereas in non-reasoning tasks (e.g., toxicity detection), thinking in non ``reasoning hub'' language (e.g., Japanese) enables LRMs to notice the safety problem and reject the user's toxic request.}
\label{fig:showcase_difference_prompt}
\end{figure}
\pagebreak 

\section{Introduction}

Recent advancements in large reasoning models (LRMs)~\citep{jaech2024openai, guo2025deepseek, muennighoff2025s1, ye2025limo} have led to striking improvements in their ability to tackle reasoning tasks such as mathematics~\cite{hendrycks2021measuring}, programming~\cite{jain2024livecodebench}, and PhD-level science questions~\cite{rein2023gpqa}.
Unlike traditional language models, LRMs employ a two-phase generation process: first, they produce a \textit{thinking} sequence where they explicitly work through intermediate reasoning steps, similar to a human's step-by-step problem-solving process. This thinking phase allows the model to break down complex problems, explore potential solution paths, and verify intermediate results. Only after completing this reasoning process does the model generate an \textit{answering} sequence that presents the final response.

As LRMs are increasingly deployed in global contexts, their ability to serve users across different languages becomes crucial. Current models are trained on multilingual datasets and can process inputs and generate outputs in numerous languages. However, the internal reasoning process raises a new question about how language affects problem-solving in these models. Our investigation reveals a striking pattern: Despite being trained on multilingual data, LRMs predominantly \textit{think} in just one or two languages, primarily English and Chinese, regardless of the input language. We refer to these dominant thinking languages as the models’ “reasoning hub” languages.

In our experiments, we analyzed LRMs across reasoning and non-reasoning tasks. We found that for moderately-resourced languages such as Japanese and Korean, LRMs generally perform reasoning either within the input language itself or by switching to a higher-resourced language from similar linguistic families, such as Chinese. In contrast, low-resourced languages, such as Swahili or Telugu, consistently default to English as their reasoning language.

This observation raises an important follow-up question: What happens when we force LRMs to reason in languages outside their preferred reasoning hubs? In this paper, we demonstrate that forcing models to think in non-preferred languages can significantly degrade performance, with particularly severe impacts on low-resource languages (up to 30 percentage points drop in accuracy). Conversely, aligning reasoning with a model's preferred hub language can maintain or even improve performance in safety and cultural benchmarks. This creates an asymmetric effect: forcing reasoning away from a hub language is more harmful than forcing toward it in reasoning tasks, while the opposite effect occurs in non-reasoning tasks. These findings have substantial implications for multilingual AI deployment.

Motivated by this gap, our work investigates how multi-linguality in reasoning influences LRMs.
Specifically, we analyze how the choice of input and reasoning languages affects LRMs from two complementary perspectives as shown in Figure \ref{fig:showcase_difference_prompt}: (1) a \textit{performance}-oriented evaluation, assessing LRMs on reasoning-intensive tasks to examine how the language used in prompting and reasoning influences their performance; and
(2) a \textit{behavior}-oriented evaluation, examining how languages impact broader aspects such as toxicity, cultural knowledge~\cite{chiu2024culturalbench}.
These aspects capture real-world implications in everyday usage scenarios.
Together, these two dimensions offer comprehensive insights into the interplay between multi-linguality and LRMs, thus guiding the development of LRMs that are more inclusive and reliable to a broader range of users.

\noindent Our contributions are as follows:

\noindent 1. We present the first comprehensive analysis of multilingual reasoning in LRMs across diverse tasks and model families. Our results demonstrate that reasoning in hub languages (English and Chinese) substantially improves accuracy on mathematical and knowledge-based tasks (by up to 26.8\%). Conversely, reasoning in non-hub languages reduces toxicity and enhances performance on cultural tasks, highlighting a critical performance–safety trade-off in multilingual AI deployment.

\noindent 2. We introduce a novel segmentation-classification method for analyzing reasoning patterns in LRMs. Using this approach, we identify systematic correlations between language-specific prefill tokens and reasoning strategies: Chinese significantly promotes subgoal setting (Pearson's r = 0.51), while English encourages backward chaining (Pearson's r = 0.30). These findings suggest that language activates distinct, culturally embedded problem-solving schemas within LRMs.




\section{Evaluation Setup}

Our evaluation framework encompasses two critical dimensions of LRM deployment: performance and behavioral alignment. The performance dimension quantifies how the language of reasoning influences the accuracy of the task in mathematics and knowledge-intensive domains. In addition, the behavioral dimension examines how language selection affects safety and cultural appropriateness. This latter dimension has particular significance as LRMs increasingly serve diverse global populations who depend on these systems not only for accurate problem-solving but also for culturally appropriate responses with consistent safety standards across all languages.


\paragraph{Reasoning Tasks}

\textbf{(i) MMMLU} extends the original MMLU~\cite{hendrycks2020measuring} test by providing human-verified translations of all 14,042 questions in 14 languages (Arabic, Bengali, German, Hindi, Japanese, Korean, Portuguese,
Russian, Spanish, Swahili, Tamil, Telugu, Thai, Yoruba). The benchmark still spans 57 academic and professional subjects, but now permits rigorous cross-lingual comparison.  We adopt the public MMMLU release\footnote{\url{https://huggingface.co/datasets/openai/MMMLU}} and its official evaluation harness. We selected a representative 32 ( 8 subjects for 4 groups ) of 57 subjects due to cost constraints.
\textbf{(ii) MATH-500} is a carefully curated subset of 500 problems from the MATH dataset \cite{hendrycks2021measuringmath}, spanning algebra, geometry, calculus, probability, and number theory. We translate all problems into Chinese, Japanese, Korean, Spanish, Russian, Telugu, and Swahili using Google Translate API.

\paragraph{Non-Reasoning Tasks}

\textbf{(i) CulturalBench} \cite{chiu2024culturalbench} evaluates models' cultural knowledge
across diverse global contexts. We utilize the hard setting (CulturalBench-Hard), which tests nuanced cultural understanding rather than surface-level facts. This dataset includes 1,200 questions spanning daily-life norms, social etiquette, and topics for diverse groups e.g., Religions across 30 countries/regions. Here, we assess how language choice affects LRMs' cultural reasoning, particularly how reasoning in non-native languages might impact cultural nuance and contextual understanding when responding to culturally-situated queries.
\textbf{(ii) LMSYS-Toxic} consists of 2,000 prompts sourced from LMSYS-1M \cite{zheng2023lmsys} that are known to trigger OpenAI's moderation API (text-moderation-latest). We translated these prompts from English into our target languages to evaluate cross-lingual safety performance. We specifically chose this dataset over alternatives such as SafetyBench \citep{zhang2023safetybench} due to its higher toxic rate, which presents a more challenging test for modern LRMs.

\subsection{Languages}
We choose English, Chinese, Spanish, Russian, Japanese, Korean, Telugu, and Swahili as the representative languages in our study.
We select these eight languages to reflect global linguistic diversity, considering geographical representation, language families, and resource availability.
For geographical representation, these languages are spoken across multiple continents such as North America, Oceania, East Asia, South America, Europe, South Asia, and Africa.
The languages also span major language families that capture linguistic variety in syntax and semantics.
Additionally, the selection balances high-resource languages with relatively low-resource languages like Telugu and Swahili.

\section{The Reasoning Hub Phenomenon in Multilingual LRMs}
\label{sec:reasoning_hub}
\begin{figure}[t!]
    \centering
    \includegraphics[width=\textwidth]{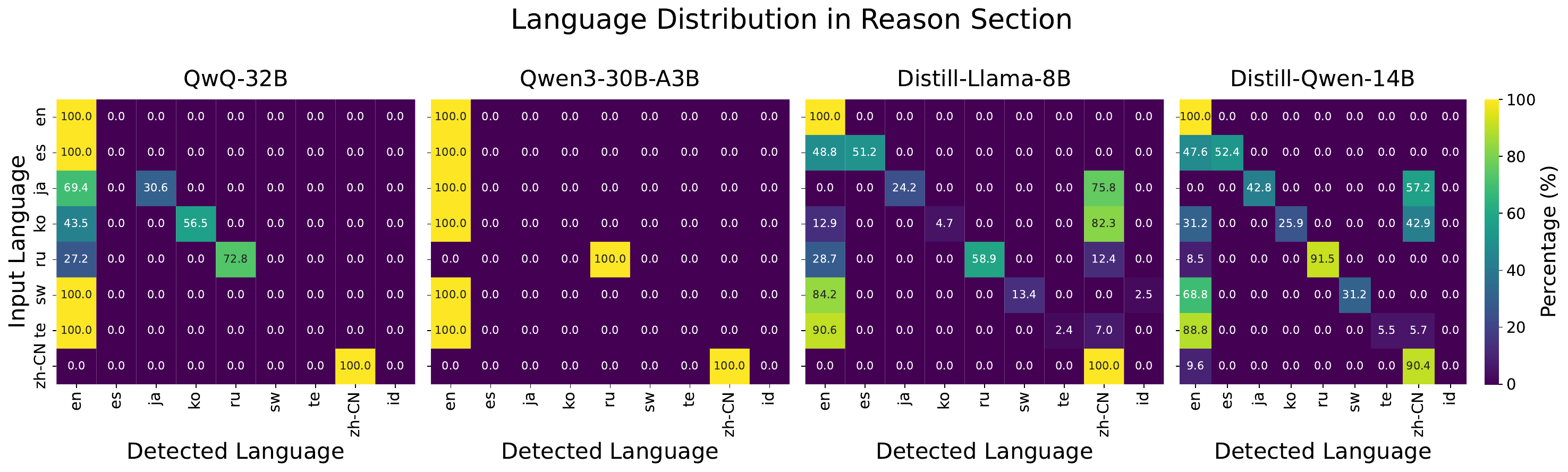}
    \vspace{0.5cm}
    \includegraphics[width=\textwidth]{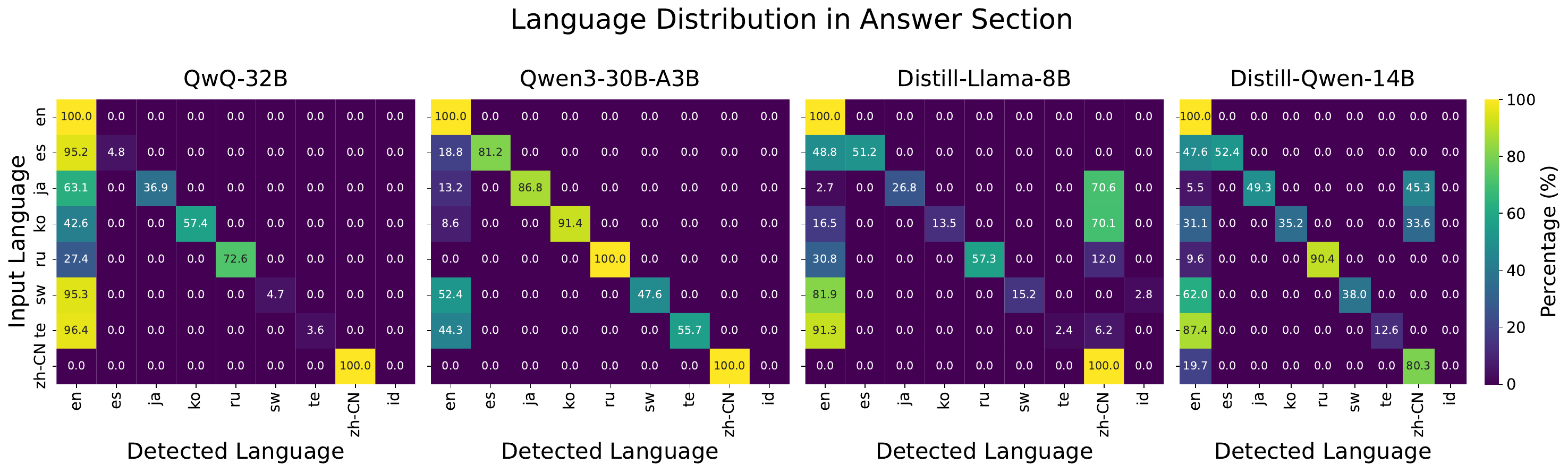}
    \vspace{0.5cm}
    \includegraphics[width=\textwidth]{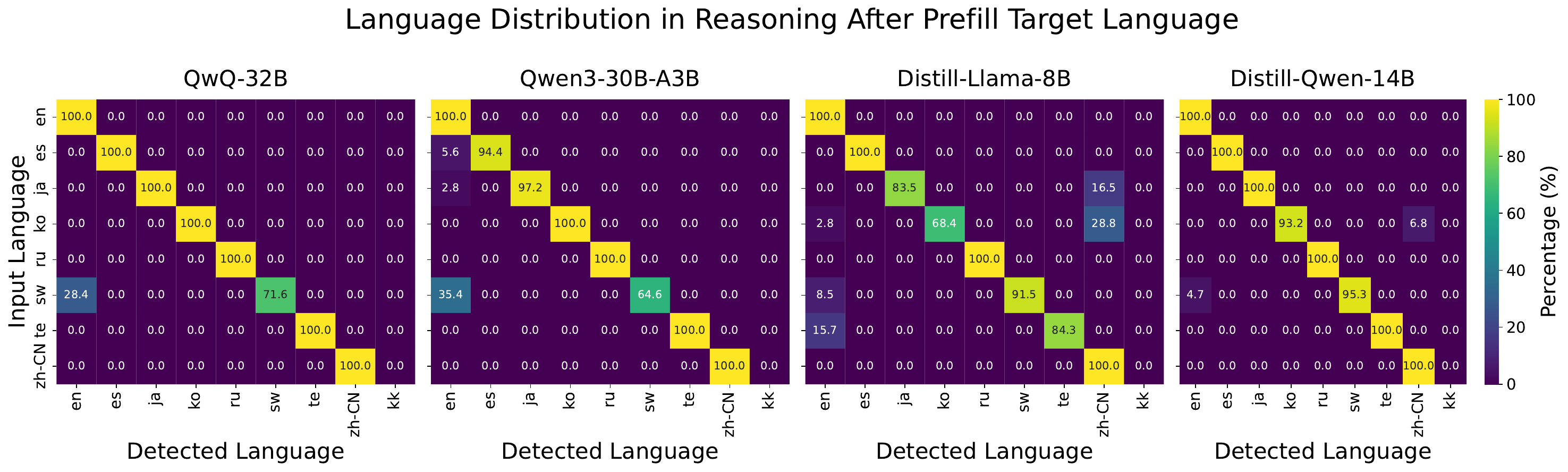}
    \caption{Language distribution visualization. \textbf{Top:} Distribution in the reason section showing language detection patterns across different models. \textbf{Middle:} Distribution in the answer section reveals how language preferences shift between reasoning and final outputs. \textbf{Bottom:} Distribution in the reason section after applying phrase prefilling, all reasoning languages were able to align well with the input language.}
    \label{fig:language_distribution_comparison}
\vskip -0.1in 
\end{figure}

While multilingual Large Language Models (LLMs) are designed to process and generate text across numerous languages, our analysis reveals a striking tendency: when generating long Chain-of-Thought (CoT) reasoning, these models predominantly default to a small subset of languages—primarily English and Chinese—regardless of the input language. We term these dominant languages ``reasoning hubs'' as they appear to function as central linguistic nodes for multilingual reasoning processes.

As illustrated in Figure \ref{fig:language_distribution_comparison}, our language detection analysis across multiple open-weight models clearly demonstrates this hub phenomenon. The top Figure shows that bigger models, such as QwQ-32B and Qwen3-30B-A3B, consistently reason in English (en) even when provided with inputs in diverse languages. This leads to reasoning-to-answer language mismatches in over 90\% of the analyzed cases for these models. 
Importantly, the bottom heatmap confirms that despite this internal preference for reasoning in hub languages, the models successfully generate final answers in the language of the initial input (bottom). This suggests a functional decoupling between the internal ``thinking'' language and the external ``responding'' language.


Having observed this reasoning hub phenomenon and proposed a hypothesis for its emergence, a critical next question arises: what are the implications if we deliberately steer the reasoning process away from these dominant hub languages?

\section{Controlling Reasoning Languages with Text Prefilling}
\label{sec:reasoning_language}

We propose a simple yet effective text prefilling strategy to steer the thinking language used by large reasoning models (LRMs) during their reasoning process, as illustrated in Figure~\ref{fig:showcase_difference_prompt}. Our method seeds the prompt with a language-specific token or phrase, following the template:

\texttt{<user> question <endoftext><assistant><think> [prefill tokens]}

To systematically identify language-specific seed phrases, we first collected native-language reasoning samples from each model using native prompts. We then extracted the first $N$ tokens (typically $T=5$–$10$) from the generated reasoning chains and computed frequency distributions over all token-level prefixes. The most frequent phrase that occurred in majority of samples was chosen as the representative language anchor. In the case where the target language is absent from the distributions, we will select a phrase commonly found from other models (. In the end, we found seed phrases such as ``Okay'' (English), ``\foreignlanguage{russian}{Хорошо}'' (Russian), ``\begin{CJK}{UTF8}{min}まず\end{CJK}'' (Japanese), ``\begin{CJK}{UTF8}{gbsn}嗯\end{CJK}'' (Chinese), ``Primero'' (Spanish), ``prārambhiṃcaḍāniki'' (Telugu) and ``Kwa'' or ``Ili kup'' (Swahili) serve as language anchors. The full prefill tokens for each model can be found in Appendix \ref{app:baseline_setting_distribution}


As demonstrated in Figure~\ref{fig:language_distribution_comparison}-Bottom, this prefilling technique substantially enhances language consistency across all evaluated models. For instance, DeepSeek-R1-Distill-Qwen-14B exhibits a much more consistent language compared to Figure \ref{fig:language_distribution_comparison}-Top, where prefilled reasoning was only partially aligned along the diagonal.

We validated our approach by comparing prefilling against token masking techniques which is less biased than our method and find no significant performance difference on MATH-500 (Appendix~\ref{app:limit_language_via_masking}). We adopted prefilling for its cross-lingual versatility with ambiguous tokenization boundaries and shared subtoken IDs.


\subsection{Performance-Oriented Results}
\label{sec:performance_oriented}
\begin{table}
\centering
\caption{Comparison of MATH-500 performance when reasoning in English vs. the target language, across languages ordered by speakers' population.}
\label{tab:math-500-prefill-comparison}
\small
\begin{tabular}{lccccccc}
\toprule
Strategy & Chinese & Spanish & Russian & Swahili & Japanese & Telugu & Korean \\
\midrule
\multicolumn{8}{c}{\textbf{DeepSeek-R1-Distill-Llama-8B}} \\
Prefill English (EN) & 78.8\% & 80.2\% & 78.4\% & 37.0\% & 74.6\% & 42.2\% & 69.8\% \\
Prefill Target Language & 73.6\% & 46.8\% & 59.4\% & 3.8\% & 32.6\% & 16.8\% & 41.6\% \\
\rowcolor{lightgreen}
Difference (EN - Target) & \textbf{+5.2\%} & \textbf{+33.4\%} & \textbf{+19.0\%} & \textbf{+33.2\%} & \textbf{+42.0\%} & \textbf{+25.4\%} & \textbf{+28.2\%} \\
\midrule
\multicolumn{8}{c}{\textbf{DeepSeek-R1-Distill-Qwen-14B}} \\
Prefill English (EN) & 88.4\% & 88.6\% & 86.6\% & 52.4\% & 85.2\% & 66.2\% & 84.4\% \\
Prefill Target Language & 89.8\% & 66.4\% & 86.4\% & 14.6\% & 63.6\% & 34.4\% & 83.8\% \\
\rowcolor{lightgreen}
Difference (EN - Target) & -1.4\% & \textbf{+22.2\%} & \textbf{+0.2\%} & \textbf{+37.8\%} & \textbf{+21.6\%} & \textbf{+31.8\%} & \textbf{+0.6\%} \\
\midrule
\multicolumn{8}{c}{\textbf{QwQ-32B}} \\
Prefill English (EN) & 92.4\% & 92.2\% & 91.2\% & 67.8\% & 90.2\% & 84.4\% & 90.6\% \\
Prefill Target Language & 90.6\% & 93.2\% & 90.6\% & 55.6\% & 87.4\% & 65.2\% & 88.2\% \\
\rowcolor{lightgreen}
Difference (EN - Target) & \textbf{+1.8\%} & -1.0\% & \textbf{+0.6\%} & \textbf{+12.2\%} & \textbf{+2.8\%} & \textbf{+19.2\%} & \textbf{+2.4\%} \\
\midrule
\multicolumn{8}{c}{\textbf{Qwen3-30B-A3B}} \\
Prefill English (EN) & 91.4\% & 91.0\% & 90.6\% & 72.4\% & 89.4\% & 87.0\% & 89.8\% \\
Prefill Target Language & 89.4\% & 83.8\% & 90.0\% & 29.6\% & 81.8\% & 68.4\% & 88.0\% \\
\rowcolor{lightgreen}
Difference (EN - Target) & \textbf{+2.0\%} & \textbf{+7.2\%} & \textbf{+0.6\%} & \textbf{+42.8\%} & \textbf{+7.6\%} & \textbf{+18.6\%} & \textbf{+1.8\%} \\
\midrule
\multicolumn{8}{c}{\textbf{Average across all models}} \\
Prefill English (EN) & 87.7\% & 88.0\% & 86.7\% & 57.4\% & 84.9\% & 70.0\% & 83.7\% \\
Prefill Target Language & 85.9\% & 72.5\% & 81.6\% & 25.9\% & 66.3\% & 46.2\% & 75.4\% \\
\rowcolor{lightgreen}
Difference (EN - Target) & \textbf{+1.9\%} & \textbf{+15.5\%} & \textbf{+5.1\%} & \textbf{+31.5\%} & \textbf{+18.5\%} & \textbf{+23.7\%} & \textbf{+8.3\%} \\
\bottomrule
\end{tabular}
\end{table}
As observed in many previous work MSGM \cite{shi2023language}, LLMs often exhibit improved performance when CoT is conducted in English, even when the primary task language is different. Our findings, presented in Table \ref{tab:math-500-prefill-comparison}, corroborate with this. Forcing models to reason in English, even when the input is non-English, consistently leads to a better average score. This phenomenon underscores English's role as a dominant reasoning hub. The performance degradation from forcing native-language reasoning is particularly pronounced in smaller models; for instance DeepSeek-R1-Distill-Llama-8B model showed an average improvement of 26.8\% with English over native reasoning. This contrasts with larger models such as Qwen-14B of 16.1\%, QwQ-32B 5.4\%, Qwen3-30B-A3B 11.5\%.

This tendency for English to serve as a more effective reasoning pathway extends beyond mathematical problem-solving, as evidenced by performance on the MMLU benchmark (Table \ref{tab:mmlu-reasoning-language-comparison}). Across various languages, employing English for reasoning steps again generally yields superior results compared to native language reasoning. This advantage is particularly striking for languages with fewer digital resources, such as Swahili, which saw improvements average of 13.6\% across all tested models. For the full models breakdown, we included it in Appendix \ref{app:mmmlu_additional}.

\begin{table}
\centering
\small
\caption{Comparison of MMLU performance when reasoning in English vs. the target language, all scores are averaged across 4 LRMs.}
\label{tab:mmlu-reasoning-language-comparison}
\begin{tabular}{lcccccc}
\toprule
Strategy & English & Chinese & Spanish & Swahili & Japanese & Korean \\
\midrule
Prefill English (EN) & -- & 83.1\% & 83.8\% & 48.8\% & 80.8\% & 77.6\% \\
Prefill Target Language & 83.0\% & 80.2\% & 78.7\% & 35.3\% & 74.0\% & 71.2\% \\
\rowcolor{lightorange}
Difference (EN - Target) & -- & \textbf{+2.9\%} & \textbf{+5.1\%} & \textbf{+13.6\%} & \textbf{+6.8\%} & \textbf{+6.4\%} \\
\bottomrule
\end{tabular}
\end{table}



\subsection{Behavior-Oriented Results}
\label{sec:behavior_oriented}

\begin{table}
    \caption{Comparison of LMSYS-Toxic ASR score when reasoning in English vs. the target language, across languages ordered by speakers' population.}
    \label{tab:toxicity}
    \centering
    \small
    \begin{tabular}{lccccccc}
    \toprule
    Strategy & Chinese & Spanish & Russian & Swahili & Japanese & Telugu & Korean \\
    \midrule
    \multicolumn{8}{c}{\textbf{Average across all models}} \\
    Prefill English (EN) & 7.7\% & 12.3\% & 11.5\% & 3.5\% & 9.9\% & 0.8\% & 4.6\% \\
    Prefill Target Language & 7.4\% & 13.3\% & 16.1\% & 3.6\% & 9.5\% & 1.6\% & 3.8\% \\
    \rowcolor{lightgreen}
    Difference (EN - Target) & +0.3\% & -1.0\% & -4.6\% & -0.1\% & +0.4\% & -0.8\% & +0.8\% \\
    \bottomrule
    \end{tabular}
    \end{table}

In LMSYS-Toxic, we observed that RL-finetuned model QwQ-32B resulted in lower attack success rate (ASR) when reasoning in their native language for most non-English languages (Japanese, Korean, Chinese, Spanish), with the notable exception of Russian. As shown in Table \ref{tab:toxicity}, QwQ-32B and Qwen3 models demonstrate a consistent pattern where forcing English reasoning (via ``Okay'' prefilling) increases toxicity rates by 1-3.5 percentage points for Japanese, Korean, Chinese, and Spanish inputs. Interestingly, the Russian language exhibits the opposite pattern, with lower toxicity when reasoning is guided toward English rather than maintaining native Russian reasoning.

This asymmetric effect aligns with our broader findings about reasoning hubs and language alignment. Both models successfully maintain the target language in their thinking phase when prompted with native language cues (>97\% native language distribution across all languages). However, the effect on toxicity varies significantly by language, suggesting that safety guardrails may be differentially effective across languages. The increased toxicity when forcing English reasoning for non-Russian languages highlights the potential safety costs of deviating from native reasoning in behavior-oriented tasks, contrasting with the performance benefits observed in the previous section.

\begin{figure}[t!] 
    \centering
        \includegraphics[width=\textwidth]{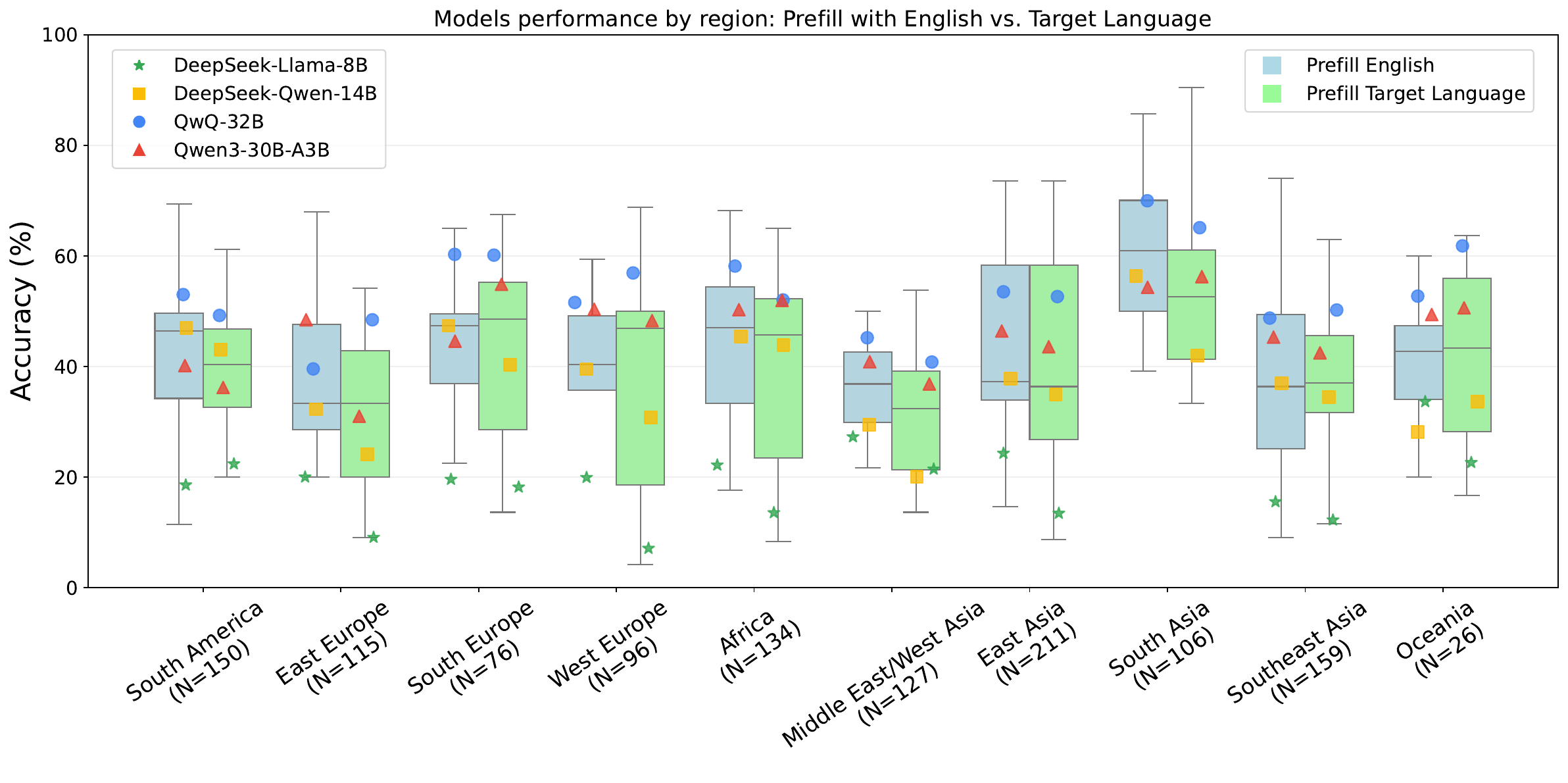}
    \caption{Model performance comparison across global regions when using English versus native language prompts}
    \label{fig:culturalbench_result}
\end{figure}
To study how changing the reasoning language affects other than safety such as culture understanding, Figure \ref{fig:culturalbench_result} compares model performance on CulturalBench-Hard (N=4907) across global regions using English versus native language. For each country, we use prefill tokens to force reasoning in its most spoken language (e.g., Nepali for Nepal, Japanese for Japan). Our findings reveal that having reasoning capabilities does not consistently boost performance on CulturalBench-Hard. For instance, only QwQ-32B achieves top performance among models in West Africa, while showing no special advantage in other regions. Having native language prompts improves CulturalBench scores in specific geo-regions, namely South Europe (+1.0\% on average) and Oceania (+2.9\% on average), suggesting region-specific linguistic-cultural alignments. 

In general, reasoning models perform best in South Asia (mean=57.3\%), similar to other non-reasoning models. Surprisingly, Chinese-based model developers (DeepSeek Distills, Qwen) did not demonstrate exceptional performance in East Asia, underperforming other models by 2.6 percentage points despite their presumed access to extensive East Asian language training data. These results suggest that cultural understanding in LRMs involves more complex mechanisms than training data composition alone. Full details can be found in Appendix \ref{app:culturalbench_details}.

Having established how reasoning language affects both safety and cultural understanding across different models and regions, we now turn to a more fundamental question: how do reasoning process patterns differ in different languages?

\section{Reasoning Pattern Analysis}
Understanding how language models reason requires a systematic analysis of their reasoning patterns. Previous approaches \citep{gandhi2025cognitive} to analyzing reasoning chains have faced two key limitations: (1) simple counting methods often overcount repeated steps, and (2) forced classification schemes assign steps to predefined categories even when they don't fit, distorting results. We propose a two-stage methodology, segmentation followed by classification, to address these limitations while enabling fine-grained analysis of reasoning behaviors.

\subsection{Segmentation-Classification Method}

\begin{figure}[t]    
    \centering
    \includegraphics[width=\textwidth]{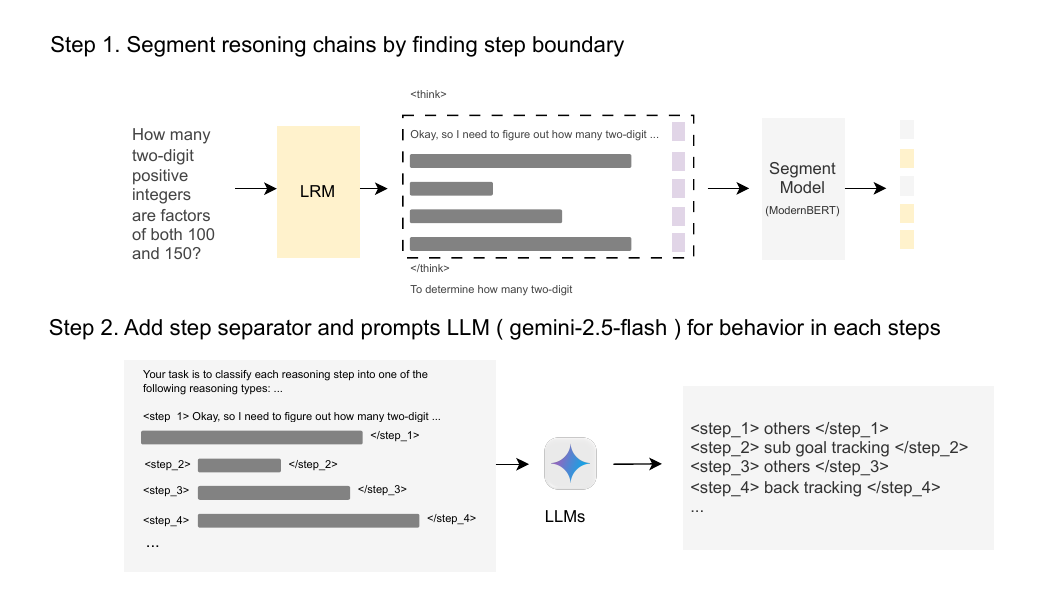}
    \caption{Two-stage pipeline for step-level category annotation of reasoning chains.}
    \label{fig:segment_classify_flow}
\end{figure}

\paragraph{Segmentation} The first stage of our methodology involves segmenting reasoning chains into distinct operational steps with clear boundaries. This segmentation is crucial for preventing overcounting and ensuring that each reasoning operation can be classified appropriately. 
We implemented a two-phase approach: (1) using GPT-4o with one-shot prompting to annotate reasoning chains by adding \texttt{<sep>} tokens between distinct operations across multilingual data from models including QwQ, Claude Sonnet, and Gemini-2.0 Flash, and (2) training a token classification model that predicts whether each sentence completes a reasoning step. We finetuned a ModernBert-large \citep{warner2024smarter}\footnote{\href{https://huggingface.co/appier-ai-research/reasoning-segmentation-model-v0}{appier-ai-research/reasoning-segmentation-model-v0}} that achieved a 95\% F1 score. Additional training details are in Appendix \ref{app:segmentation_details}.

\paragraph{Classification} The second stage involves classifying each segmented step according to a theoretically grounded taxonomy. Building upon the four habits \citep{gandhi2025cognitive} taxonomy, we examine four primary habits that have demonstrated empirical significance in LRMs. 
We classify each segmented reasoning step using gemini-2.0-flash, according to four primary cognitive habits from \citep{gandhi2025cognitive}:

\begin{compactitem}

    \item \textbf{Subgoal setting:} Where the model breaks down the problem into smaller, intermediate goals (e.g., "To solve this, we first need to..." or "First, I'll try to..., then...").
    
    \item \textbf{Backtracking:} Where the model realizes a path won't work and explicitly goes back to try a different approach (e.g., "Let me try again" or "We need to try a different approach").
    
    \item \textbf{Verification:} Where the model checks the correctness of intermediate results or ensures the final answer is correct (e.g., "Let's verify this calculation" or "Checking our solution...").
    
    \item \textbf{Backward chaining:} Where the model works backward from its answer to see whether it can derive the variables in the original problem (e.g., "If we want to reach 42, then we need...").
\end{compactitem}

To address the inherent limitations of fixed taxonomies, we introduce an ``Others'' category for steps that don't clearly fit the defined habits, preventing distortion from forced classification. This category allows us to identify true novel reasoning behaviors or variations, ensuring we do not over-count while acknowledging the diversity of reasoning strategies across models. Full prompts can be found in Appendix \ref{app:classify_reasoning_method}.

\subsection{Reasoning Behaviours and Performance across Models and Linguistic Contexts} 

\begin{figure}[t!]
\centering
\begin{minipage}[t]{0.49\textwidth}
    \centering
    \includegraphics[width=\linewidth]{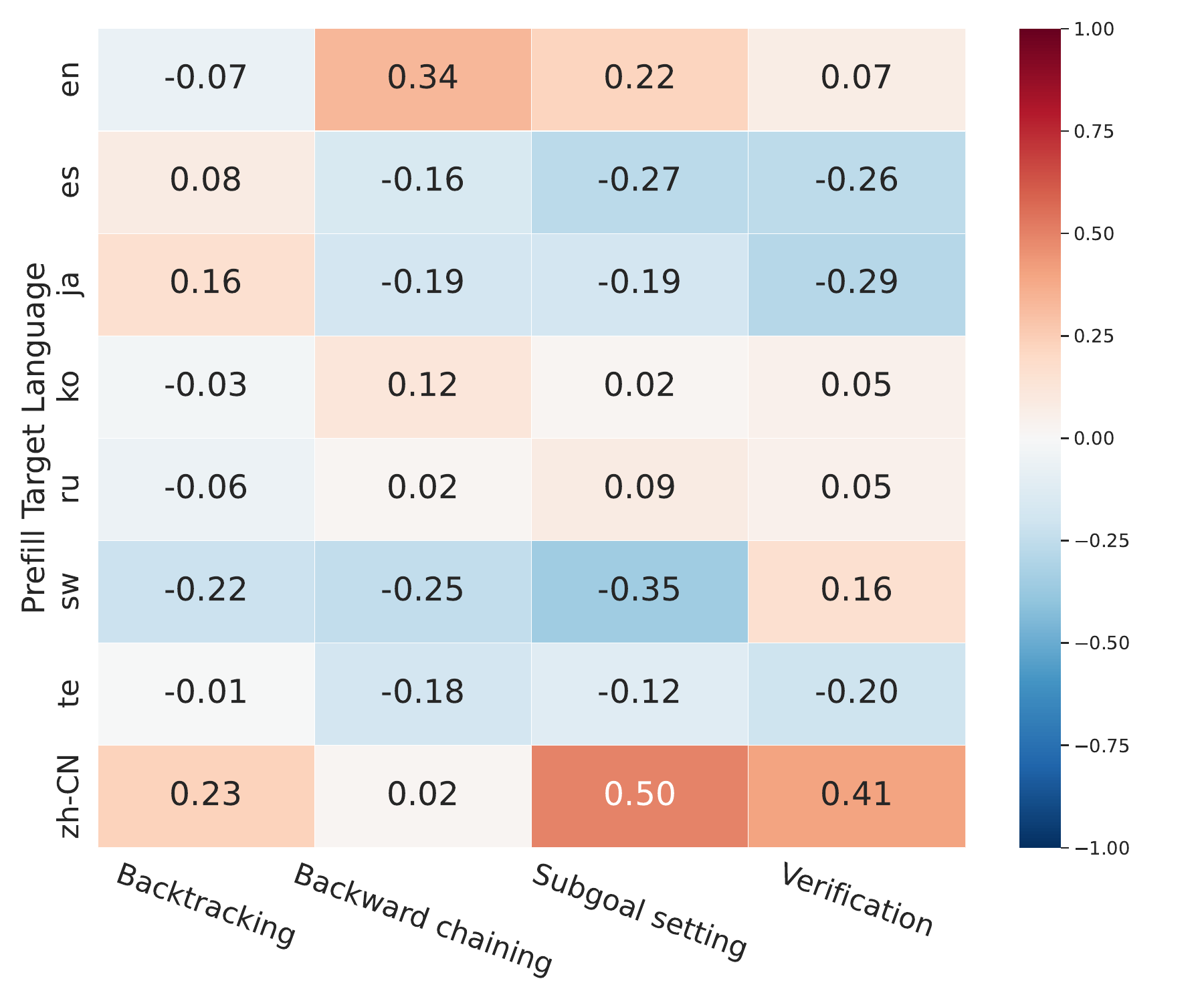}
    \caption{Correlation Matrix Between Prefill Target Languages and Reasoning Types}
    \label{fig:prefill_correlation_matrix}
\end{minipage}
\hfill
\begin{minipage}[t]{0.49\textwidth}
    \centering
    \includegraphics[width=\linewidth]{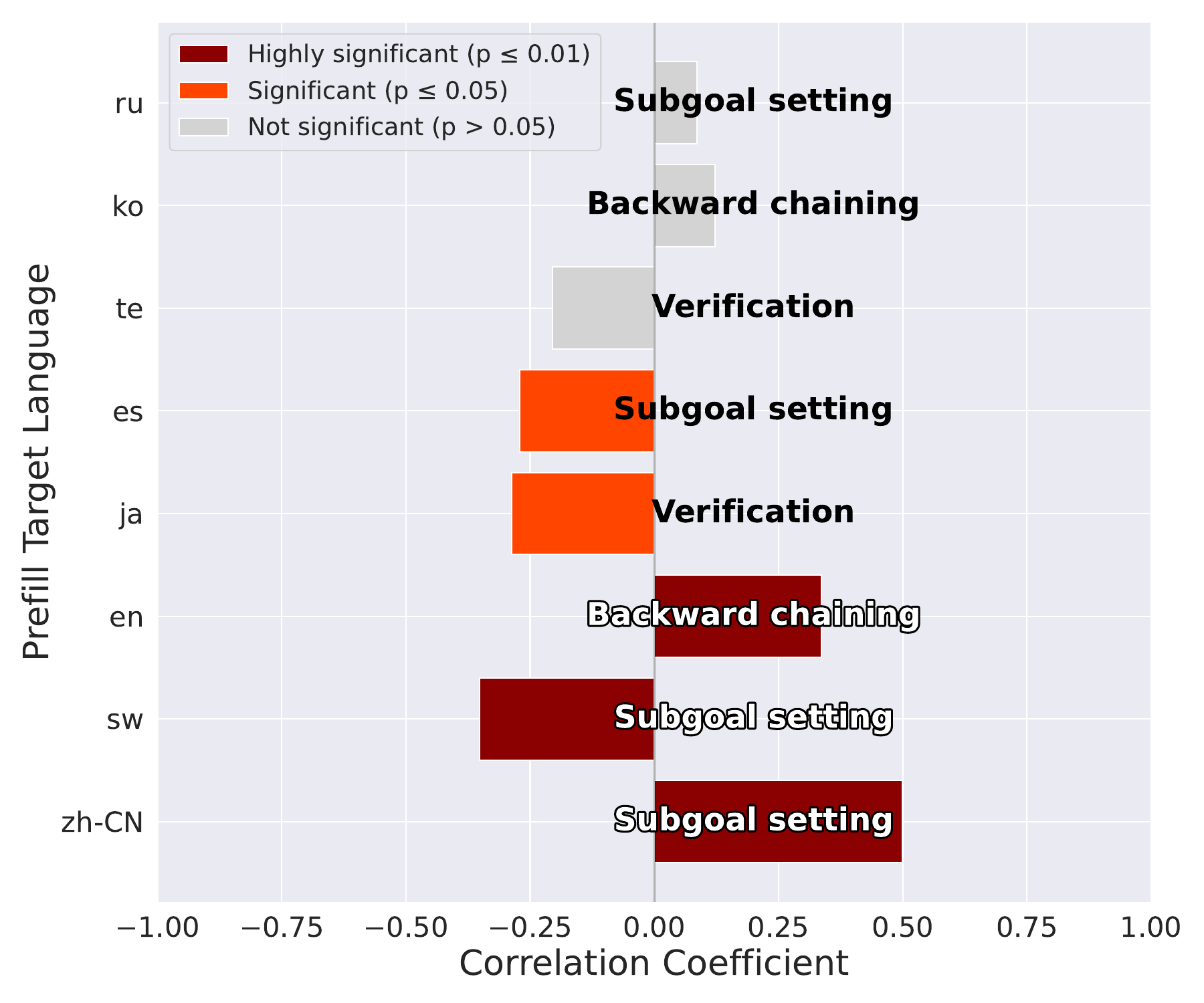}
    \caption{Figure shows the behavior with the strongest correlation for each language. Bar colors indicate statistical significance levels.}
    \label{fig:behavior_relevant}
\end{minipage}
\vskip -0.2in 
\end{figure}

To analyze how prefill target languages affect specific reasoning strategies, we computed Pearson correlation coefficients (r) and their corresponding p-values (p). The process involved first aggregating the experimental results from all four models. For each experimental setting—defined by a unique combination of input language and prefill target language—we calculated the average count of steps for each of the four reasoning habits. Subsequently, for each prefill target language (e.g., Chinese, Swahili) and each reasoning habit (e.g., Subgoal setting), we calculate the pearson correlations between the average counts per reasoning and the final accuracy.

Figure~\ref{fig:prefill_correlation_matrix} demonstrates that these minimal linguistic cues fundamentally reshape reasoning approaches—Chinese prefill tokens strongly promote Subgoal setting (r=0.50, p<0.001) and Verification (r=0.41, p<0.001), while Swahili shows a significant negative correlation (r=-0.35, p<0.01) with the same Subgoal setting behavior.

We hypothesize this effect stems from culturally embedded problem-solving schemas activated by language-specific tokens. This aligns with cognitive linguistic structures that prime different decomposition strategies, as documented in bilingual problem-solving studies\citep{bernardo2005effects}.

The distinct reasoning ``signatures'' in Figure~\ref{fig:behavior_relevant} further support this hypothesis—English uniquely encourages Backward chaining (r=0.30, p<0.015), a deductive approach consistent with Anglo-Saxon educational emphases on proof-based reasoning. These signatures persist across model architectures, suggesting we're observing fundamental interactions between language and cognition rather than model-specific artifacts.
Performance analysis reveals that models employing Subgoal setting strategies (predominantly triggered by Chinese prefilings) achieved 7.3\% higher accuracy on MATH-500 problems compared to those using other dominant strategies.
This suggests that by strategically selecting prefill languages, we can optimize model performance on tasks requiring specific reasoning approaches.

\section{Related Work}

\subsection{Chain-of-Thought Analysis}

Chain-of-thought (CoT) prompting enhances large language models' reasoning capabilities by generating explicit intermediate steps, improving performance, and providing interpretable insights into decision processes. Resources such as \textit{ThoughtSource} ~\citep{ott2023thoughtsource} support systematic CoT evaluation across diverse domains, but recent evidence shows that the verbalized chains of the models are not always faithful by \cite{chen2025reasoning}, which shows that reasoning models can omit crucial shortcuts (e.g., hidden hints or implicit translations—suggesting a misalignment between the true internal process and the stated CoT). Complementary analysis by \citep{opielka2025analogical} indicates that LLMs reuse reasoning patterns through ``concept vectors'' encoding structural relationships consistently across tasks, implying that models map new problems to analogously solved ones through shared building blocks.



\subsection{Hub Languages and Reasoning in Multilingual LLMs}
The concept of a ``hub language'' facilitating cross-lingual understanding originated in information retrieval, where ~\cite{rupnik2012cross} showed how resource-rich languages like English could bridge document retrieval between language pairs lacking direct comparable corpora. Building on this, ~\citep{wu2024semantic} proposed the ``Semantic Hub Hypothesis'', suggesting LLMs develop a shared representation space across languages, with the model's dominant pretraining language (typically English) scaffolding this hub and influencing outputs in other languages. Further evidence from ~\citep{schut2025multilingual} demonstrates, through logit lens analysis, that non-English inputs are often processed via English-aligned representations in intermediate layers before translation back to the input language. Behaviorally, ~\citep{etxaniz2024multilingual} found LLMs achieve superior performance when non-English inputs are first translated to English for processing. These findings suggest many LLMs default to an English-centric reasoning pathway internally despite their multilingual capabilities. Our research contributes to this discussion by systematically analyzing reasoning in LRMs (Section~\ref{sec:reasoning_hub}) and the impact of forcing reasoning in specific languages (Sections~\ref{sec:performance_oriented} and \ref{sec:behavior_oriented}).

\section{Conclusion}

In this work, we reveal that LRMs, despite their strong multilingual ability, predominantly still prefer to reason in hub languages such as English, regardless of the input language. Our introduction of a text pre-filling method provides a practical approach to guide the reasoning language with high success. We demonstrated an asymmetric effect: forcing models to reason in non-hub languages degrades performance in low-resource languages, whereas aligning reasoning with hub languages improves or maintains the performance in reasoning tasks. However, in the cultural reasoning task, native-language reasoning can be beneficial. These findings underscore the critical importance of considering the internal reasoning language to be more inclusive for future models.

\section{Limitations}
\label{sec:limitation}

In our work, we only study eight languages, which may not fully represent the diversity of global languages, particularly extremely low-resource ones. Our reasoning-analysis pipeline depends on LLM annotators and a relatively coarse four-habit taxonomy, which may mask subtler reasoning strategies that differ across languages. While we identify significant correlations between languages and reasoning approaches, we cannot establish causal relationships without more controlled experiments. Additionally, our analysis is limited to medium-scale LRMs (< 30B), and the reasoning hub phenomenon may evolve as model scales. 

\bibliographystyle{plainnat}
\bibliography{references.bib}

\begin{thebibliography}{24}
\providecommand{\natexlab}[1]{#1}
\providecommand{\url}[1]{\texttt{#1}}
\expandafter\ifx\csname urlstyle\endcsname\relax
  \providecommand{\doi}[1]{doi: #1}\else
  \providecommand{\doi}{doi: \begingroup \urlstyle{rm}\Url}\fi

\bibitem[Bernardo and Calleja(2005)]{bernardo2005effects}
Allan~BI Bernardo and Marissa~O Calleja.
\newblock The effects of stating problems in bilingual students' first and second languages on solving mathematical word problems.
\newblock \emph{The Journal of Genetic Psychology}, 166\penalty0 (1):\penalty0 117--129, 2005.

\bibitem[Chen et~al.(2025)Chen, Benton, Radhakrishnan, Uesato, Denison, Schulman, Somani, Hase, Wagner, Roger, et~al.]{chen2025reasoning}
Yanda Chen, Joe Benton, Ansh Radhakrishnan, Jonathan Uesato, Carson Denison, John Schulman, Arushi Somani, Peter Hase, Misha Wagner, Fabien Roger, et~al.
\newblock Reasoning models don't always say what they think.
\newblock \emph{arXiv preprint arXiv:2505.05410}, 2025.

\bibitem[Chiu et~al.(2024)Chiu, Jiang, Lin, Park, Li, Ravi, Bhatia, Antoniak, Tsvetkov, Shwartz, et~al.]{chiu2024culturalbench}
Yu~Ying Chiu, Liwei Jiang, Bill~Yuchen Lin, Chan~Young Park, Shuyue~Stella Li, Sahithya Ravi, Mehar Bhatia, Maria Antoniak, Yulia Tsvetkov, Vered Shwartz, et~al.
\newblock Culturalbench: a robust, diverse and challenging benchmark on measuring the (lack of) cultural knowledge of llms.
\newblock \emph{arXiv preprint arXiv:2410.02677}, 2024.

\bibitem[Etxaniz et~al.(2024)Etxaniz, Azkune, Soroa, Lacalle, and Artetxe]{etxaniz2024multilingual}
Julen Etxaniz, Gorka Azkune, Aitor Soroa, Oier Lacalle, and Mikel Artetxe.
\newblock Do multilingual language models think better in english?
\newblock In \emph{Proceedings of the 2024 Conference of the North American Chapter of the Association for Computational Linguistics: Human Language Technologies (Volume 2: Short Papers)}, pages 550--564, 2024.

\bibitem[Gandhi et~al.(2025)Gandhi, Chakravarthy, Singh, Lile, and Goodman]{gandhi2025cognitive}
Kanishk Gandhi, Ayush Chakravarthy, Anikait Singh, Nathan Lile, and Noah~D Goodman.
\newblock Cognitive behaviors that enable self-improving reasoners, or, four habits of highly effective stars.
\newblock \emph{arXiv preprint arXiv:2503.01307}, 2025.

\bibitem[Guo et~al.(2025)Guo, Yang, Zhang, Song, Zhang, Xu, Zhu, Ma, Wang, Bi, et~al.]{guo2025deepseek}
Daya Guo, Dejian Yang, Haowei Zhang, Junxiao Song, Ruoyu Zhang, Runxin Xu, Qihao Zhu, Shirong Ma, Peiyi Wang, Xiao Bi, et~al.
\newblock Deepseek-r1: Incentivizing reasoning capability in llms via reinforcement learning.
\newblock \emph{arXiv preprint arXiv:2501.12948}, 2025.

\bibitem[Hendrycks et~al.(2020)Hendrycks, Burns, Basart, Zou, Mazeika, Song, and Steinhardt]{hendrycks2020measuring}
Dan Hendrycks, Collin Burns, Steven Basart, Andy Zou, Mantas Mazeika, Dawn Song, and Jacob Steinhardt.
\newblock Measuring massive multitask language understanding.
\newblock \emph{arXiv preprint arXiv:2009.03300}, 2020.

\bibitem[Hendrycks et~al.(2021{\natexlab{a}})Hendrycks, Burns, Kadavath, Arora, Basart, Tang, Song, and Steinhardt]{hendrycks2021measuring}
Dan Hendrycks, Collin Burns, Saurav Kadavath, Akul Arora, Steven Basart, Eric Tang, Dawn Song, and Jacob Steinhardt.
\newblock Measuring mathematical problem solving with the math dataset.
\newblock \emph{arXiv preprint arXiv:2103.03874}, 2021{\natexlab{a}}.

\bibitem[Hendrycks et~al.(2021{\natexlab{b}})Hendrycks, Burns, Kadavath, Arora, Basart, Tang, Song, and Steinhardt]{hendrycks2021measuringmath}
Dan Hendrycks, Collin Burns, Saurav Kadavath, Akul Arora, Steven Basart, Eric Tang, Dawn Song, and Jacob Steinhardt.
\newblock Measuring mathematical problem solving with the math dataset.
\newblock \emph{arXiv preprint arXiv:2103.03874}, 2021{\natexlab{b}}.

\bibitem[Jaech et~al.(2024)Jaech, Kalai, Lerer, Richardson, El-Kishky, Low, Helyar, Madry, Beutel, Carney, et~al.]{jaech2024openai}
Aaron Jaech, Adam Kalai, Adam Lerer, Adam Richardson, Ahmed El-Kishky, Aiden Low, Alec Helyar, Aleksander Madry, Alex Beutel, Alex Carney, et~al.
\newblock Openai o1 system card.
\newblock \emph{arXiv preprint arXiv:2412.16720}, 2024.

\bibitem[Jain et~al.(2024)Jain, Han, Gu, Li, Yan, Zhang, Wang, Solar-Lezama, Sen, and Stoica]{jain2024livecodebench}
Naman Jain, King Han, Alex Gu, Wen-Ding Li, Fanjia Yan, Tianjun Zhang, Sida Wang, Armando Solar-Lezama, Koushik Sen, and Ion Stoica.
\newblock Livecodebench: Holistic and contamination free evaluation of large language models for code.
\newblock \emph{arXiv preprint arXiv:2403.07974}, 2024.

\bibitem[Muennighoff et~al.(2025)Muennighoff, Yang, Shi, Li, Fei-Fei, Hajishirzi, Zettlemoyer, Liang, Cand{\`e}s, and Hashimoto]{muennighoff2025s1}
Niklas Muennighoff, Zitong Yang, Weijia Shi, Xiang~Lisa Li, Li~Fei-Fei, Hannaneh Hajishirzi, Luke Zettlemoyer, Percy Liang, Emmanuel Cand{\`e}s, and Tatsunori Hashimoto.
\newblock s1: Simple test-time scaling.
\newblock \emph{arXiv preprint arXiv:2501.19393}, 2025.

\bibitem[Opie{\l}ka et~al.(2025)Opie{\l}ka, Rosenbusch, and Stevenson]{opielka2025analogical}
Gustaw Opie{\l}ka, Hannes Rosenbusch, and Claire~E Stevenson.
\newblock Analogical reasoning inside large language models: Concept vectors and the limits of abstraction.
\newblock \emph{arXiv preprint arXiv:2503.03666}, 2025.

\bibitem[Ott et~al.(2023)Ott, Hebenstreit, Li{\'e}vin, Hother, Moradi, Mayrhauser, Praas, Winther, and Samwald]{ott2023thoughtsource}
Simon Ott, Konstantin Hebenstreit, Valentin Li{\'e}vin, Christoffer~Egeberg Hother, Milad Moradi, Maximilian Mayrhauser, Robert Praas, Ole Winther, and Matthias Samwald.
\newblock Thoughtsource: A central hub for large language model reasoning data.
\newblock \emph{Scientific data}, 10\penalty0 (1):\penalty0 528, 2023.

\bibitem[Rein et~al.(2023)Rein, Hou, Stickland, Petty, Pang, Dirani, Michael, and Bowman]{rein2023gpqa}
David Rein, Betty~Li Hou, Asa~Cooper Stickland, Jackson Petty, Richard~Yuanzhe Pang, Julien Dirani, Julian Michael, and Samuel~R Bowman.
\newblock Gpqa: A graduate-level google-proof q\&a benchmark.
\newblock \emph{arXiv preprint arXiv:2311.12022}, 2023.

\bibitem[Rupnik et~al.(2012)Rupnik, Muhic, and Skraba]{rupnik2012cross}
Jan Rupnik, Andrej Muhic, and P~Skraba.
\newblock Cross-lingual document retrieval through hub languages.
\newblock In \emph{Neural Information Processing Systems Workshop}, 2012.

\bibitem[Schut et~al.(2025)Schut, Gal, and Farquhar]{schut2025multilingual}
Lisa Schut, Yarin Gal, and Sebastian Farquhar.
\newblock Do multilingual llms think in english?
\newblock \emph{arXiv preprint arXiv:2502.15603}, 2025.

\bibitem[Shi et~al.(2023)Shi, Suzgun, Freitag, Wang, Srivats, Vosoughi, Chung, Tay, Ruder, Zhou, et~al.]{shi2023language}
Freda Shi, Mirac Suzgun, Markus Freitag, Xuezhi Wang, Suraj Srivats, Soroush Vosoughi, Hyung~Won Chung, Yi~Tay, Sebastian Ruder, Denny Zhou, et~al.
\newblock Language models are multilingual chain-of-thought reasoners.
\newblock In \emph{The Eleventh International Conference on Learning Representations}, 2023.

\bibitem[Warner et~al.(2024)Warner, Chaffin, Clavi{\'e}, Weller, Hallstr{\"o}m, Taghadouini, Gallagher, Biswas, Ladhak, Aarsen, et~al.]{warner2024smarter}
Benjamin Warner, Antoine Chaffin, Benjamin Clavi{\'e}, Orion Weller, Oskar Hallstr{\"o}m, Said Taghadouini, Alexis Gallagher, Raja Biswas, Faisal Ladhak, Tom Aarsen, et~al.
\newblock Smarter, better, faster, longer: A modern bidirectional encoder for fast, memory efficient, and long context finetuning and inference.
\newblock \emph{arXiv preprint arXiv:2412.13663}, 2024.

\bibitem[Wu et~al.(2024)Wu, Yu, Yogatama, Lu, and Kim]{wu2024semantic}
Zhaofeng Wu, Xinyan~Velocity Yu, Dani Yogatama, Jiasen Lu, and Yoon Kim.
\newblock The semantic hub hypothesis: Language models share semantic representations across languages and modalities.
\newblock \emph{arXiv preprint arXiv:2411.04986}, 2024.

\bibitem[Ye et~al.(2025)Ye, Huang, Xiao, Chern, Xia, and Liu]{ye2025limo}
Yixin Ye, Zhen Huang, Yang Xiao, Ethan Chern, Shijie Xia, and Pengfei Liu.
\newblock Limo: Less is more for reasoning.
\newblock \emph{arXiv preprint arXiv:2502.03387}, 2025.

\bibitem[Yu et~al.(2023)Yu, Jiang, Shi, Yu, Liu, Zhang, Kwok, Li, Weller, and Liu]{yu2023metamath}
Longhui Yu, Weisen Jiang, Han Shi, Jincheng Yu, Zhengying Liu, Yu~Zhang, James~T Kwok, Zhenguo Li, Adrian Weller, and Weiyang Liu.
\newblock Metamath: Bootstrap your own mathematical questions for large language models.
\newblock \emph{arXiv preprint arXiv:2309.12284}, 2023.

\bibitem[Zhang et~al.(2023)Zhang, Lei, Wu, Sun, Huang, Long, Liu, Lei, Tang, and Huang]{zhang2023safetybench}
Zhexin Zhang, Leqi Lei, Lindong Wu, Rui Sun, Yongkang Huang, Chong Long, Xiao Liu, Xuanyu Lei, Jie Tang, and Minlie Huang.
\newblock Safetybench: Evaluating the safety of large language models with multiple choice questions.
\newblock \emph{arXiv preprint arXiv:2309.07045}, 2023.

\bibitem[Zheng et~al.(2023)Zheng, Chiang, Sheng, Li, Zhuang, Wu, Zhuang, Li, Lin, Xing, et~al.]{zheng2023lmsys}
Lianmin Zheng, Wei-Lin Chiang, Ying Sheng, Tianle Li, Siyuan Zhuang, Zhanghao Wu, Yonghao Zhuang, Zhuohan Li, Zi~Lin, Eric~P Xing, et~al.
\newblock Lmsys-chat-1m: A large-scale real-world llm conversation dataset.
\newblock \emph{arXiv preprint arXiv:2309.11998}, 2023.

\end{thebibliography}

\newpage 

\appendix 
\part*{Appendices}
\section{Model Details}
\label{app:model_details}

We use the latest sglang inference engine to evaluate all open weights model on A100 GPU with the exception of QwQ-32B which uses Together.ai serverless API endpoint.

As of the decoding parameters we used for all models which was recommended by original model provider Table \ref{tab:decoding-params}.

\begin{table}[t]
  \centering
  \small
  \caption{Decoding parameters used for each model during evaluation.}
  \label{tab:decoding-params}
  \begin{tabular}{lcccc}
    \toprule
    \textbf{Model} & \textbf{Temperature} & \textbf{Top-\textit{p}} & \textbf{Top-\textit{k}} & \textbf{Min-\textit{p}} \\
    \midrule
    DeepSeek-R1-Distill-Llama-8B           & 0.6 & 0.95 & --- & --- \\
    DeepSeek-R1-Distill-Qwen-14B           & 0.6 & 0.95 & --- & --- \\
    Qwen3-30B-A3B (reasoning on / off)     & 0.6 & 0.95 & 20  & 0   \\
    QwQ-32B                                & 0.6 & 0.95 & --- & 0   \\
    \bottomrule
  \end{tabular}
\end{table}

For the base model experiments found in Table \ref{tab:aime-base-zero-prompt-scores}, we simply set temperature = 0.6 only.

\subsection{Inference Cost}
\label{app:inference_cost_breakdown}

QwQ-32B cost around 600 USD for all the experiments including ablation studies in scaling efficiency. While other models: Deepseek-Distill-Qwen-14B, Deepseek-Distill-Llama-8B, Qwen3-30B-A3B cost around 1,200 USD in A100 GPUs cost calculated at 1.8 USD per hour per card. 

The entire inference process took over 2 weeks to finish under 2 A100 GPUs, using the latest sglang inference service.


\clearpage
\section{Reasoning Process Analysis}
\label{app:reasoning_analysis_details}
\subsection{Segmentation Details}
\label{app:segmentation_details}

In this section we provided the details we used to curate dataset and the training our segmentation model.

\paragraph{Dataset} We collect existing reasoning dataset shared by others from huggingface. We mainly collect reasoning process from Deepseek-R1, Deepseek-R1-Zero, Gemini-2.0-Flash, Claude-3-7-Sonnet, QwQ-preview, MetaMath CoT response \citep{yu2023metamath} and Open-R1 : an attempt to generate long CoT from Qwen models. The amount of reasonings from each models can be found in Table \ref{tab:model_data_counts}. For each reasoning, we prompt gpt-4o-2024-07-18 with 1-shot segmentation prompt to segment the reasoning text into steps. Prompts can be found in Figure \ref{fig:segmentation_prompt}. The raw output is then processed into a sequence chunk which we can used to train a small segmentation model. The annotation cost around 35 USD without any batch discount.

\begin{table}[t]
\caption{Data Count Distribution Across Models}
\label{tab:model_data_counts}
\centering
\begin{tabular}{lc}
\toprule
\textbf{Model} & \textbf{Count} \\
\midrule
deepseek-r1-zero & 647 \\
meta math & 539 \\
gemini-flash-thinking & 530 \\
deepseek-r1 & 517 \\
qwq-preview & 506 \\
metamath-qwen & 402 \\
openr1-preview & 116 \\
claude-3-7 & 47 \\
\bottomrule
\end{tabular}
\end{table}

\paragraph{Hyperparameters} We split the dataset into 7:3 train and validation set. And we simply use the validation to select the best hyperparameters as found in Table \ref{tab:training_params} which achieve a high F1 score of 96.08. Training a single hyperparameters took around 4 hours to finished on 4090 GPU.

\begin{figure}[t]    \centering
        \includegraphics[width=\textwidth]{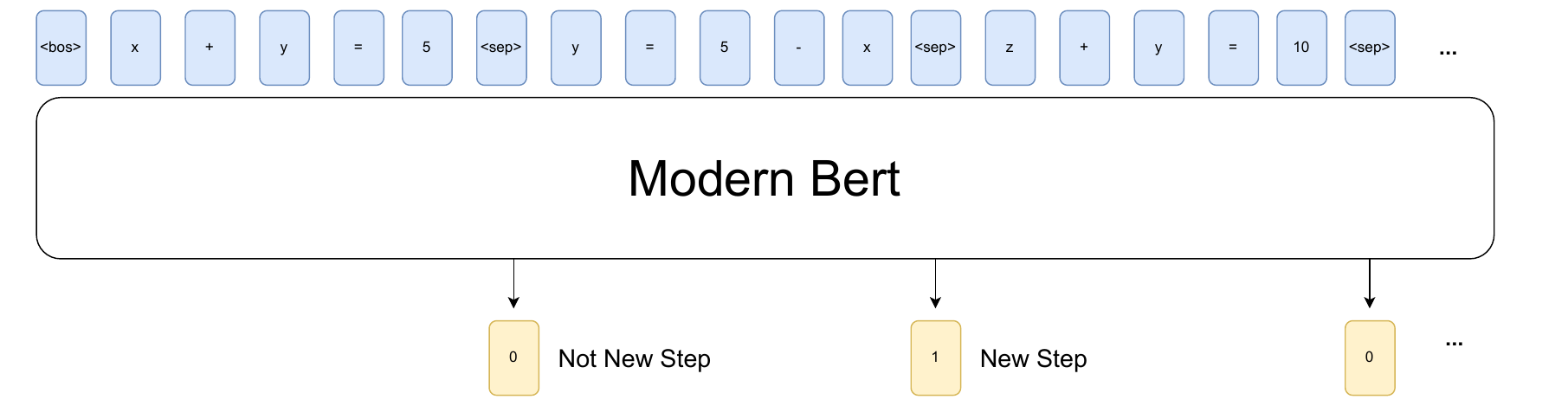}
    \caption{A showcase of how segmentation prediction works }
    \label{fig:modern_bert_segmentation}
\end{figure}

\paragraph{Inputs and Target Formats} Figure \ref{fig:modern_bert_segmentation} illustrates the ModernBERT segmentation process. For each thinking process extracted from model responses, we first split the text by newline symbols, replacing each with a special token (<sep>). The model is trained to predict whether each <sep> token indicates the beginning of a new reasoning step (1) or the continuation of the current step (0). As shown in the figure, ModernBERT takes a reasoning sequence as input (top) and processes mathematical expressions (x + y = 5, y = 5 - x, z + y = 10), classifying each separator position to enable structured parsing of complex reasoning chains. This binary classification approach allows the model to effectively identify logical breakpoints in reasoning processes.

\begin{table}[t]
\caption{Training Parameters for ModernBERT-large}
\label{tab:training_params}
\centering
\begin{tabular}{lcc}
\toprule
\textbf{Parameter} & \textbf{Best} & \textbf{Searched} \\
\midrule
Learning Rate & $8 \times 10^{-5}$ & $\{5 \times 10^{-5}, 8 \times 10^{-5}, 1 \times 10^{-4}, 3 \times 10^{-4}\}$ \\
Batch Size & 24 & \{16, 24, 32\} \\
Weight Decay & 0.01 & - \\
Number of Epochs & 10 & - \\
Warmup Steps & 50 & - \\
Optimizer & AdamW & - \\
\bottomrule
\end{tabular}
\end{table}

\clearpage    

\begin{minipage}{\textwidth}
    \begin{mdframed}

   Output your segmentation result by adding a <sep> to the original text to indicate a separation between steps\\
\\
    Do not modify the original reasoning text, only add a separation token\\   
\\
    Do not split table into segments, keep a whole table as one step\\
\\
\\
    \# Example\\
\\
    \texttt{[INPUT]:}\\
    ```\\
    Okay, let's see. ...\\
    Alright, let's break this down. First, ...\\
    \\
    **Final Answer**\\
    \verb|\boxed{251.60}|\\
    ```\\
\\
    \texttt{[OUTPUT]:}\\
    ```\\
    Okay, let's see. So ...\\
    <sep>\\
    Alright, let's break this down. ...\\
    \texttt{[Skip for brevity]}\\
    ...\\
    <sep>\\
    **Final Answer**\\
    \verb|\boxed{251.60}|\\
    ```\\
\\
    Now do the same task by following the same pattern as above:\\
\\
    \texttt{[INPUT]:}\\
    ```\\
    \texttt{{thinking process goes here}}\\
    ```\\
\\
    \texttt{[OUTPUT]:}
    \end{mdframed}
    \captionsetup{hypcap=false}
    \captionof{figure}{The prompt template for segmenting reasoning steps with <sep> tokens.}
    \label{fig:segmentation_prompt}
\end{minipage}

\subsection{Reasoning Process Classification}
\label{app:classify_reasoning_method}
After segmentation, we concatenate the individual reasoning processes using numbered step tokens (e.g., \textless step\_1\textgreater reasoning process 1 \textless step\_1\textgreater \textbackslash n \textless step\_2\textgreater reasoning process 2 \textless step\_2\textgreater ...). This structured sequence, along with the original question, is then passed to a classification prompt as illustrated in Figure \ref{fig:classify_prompt}. We utilize gemini-2.0-flash to perform the classification of each reasoning step according to our taxonomy.

While we initially explored more sophisticated taxonomies that included problem reading and abduction classification, the complexity of these frameworks exceeded the classification capabilities of current LLMs, limiting potential downstream insights. We therefore opted for the simpler four-habits taxonomy. Investigating more complex taxonomies remains an avenue for future research.

\begin{minipage}{\textwidth}
    \begin{mdframed}
        Here is a problem and the reasoning process that an LLM generated when it tries to solve the problem.\\
    
        Problem: (enclosed in double backticks)\\
        ``\\
        {problem}\\
        ``\\
    
        Reasoning process: (enclosed in triple backticks, the reasoning process has been split into distinct reasoning steps in the format of \textless step\_idx\textgreater\textless reasoning\_step\_content\textgreater\textless /step\_idx\textgreater)\\
        ```\\
        {reasoning}\\
        ```\\
    
        Your task is to classify each reasoning step into one of the following reasoning types: (specified by \textless type\_index\textgreater. \textless type\_name\textgreater: \textless definition\textgreater)

        1. Subgoal setting: Where the model breaks down the problem into smaller, intermediate goals (e.g., 'To solve this, we first need to...' or 'First, I'll try to ..., then ...'\\
        2. Backtracking: Where the model realizes a path won't work and explicitly goes back to try a different approach. An example of backtracking is: 'Let me try again' or 'we need to try a different approach'.\\
        3. Verification: Where the model checks the correctness of the intermediate results or to make sure the final answer is correct.\\
        4. Backward chaining: Where the model works backward from its answer to see whether it can derive the variables in the original problem.\\
        5. Others: This reasoning step is the continuation of the previous reasoning step, or it does not fall into any of the above categories.\\
    
        Generate the rationale before you make the classification.
        Provide your output in the following format:\\
    
        [Reasoning]\\
        
        \textless step\_1\textgreater\textless rationale\_1\textgreater\textless type\_name\_1\textgreater\textless /step\_1\textgreater \\
        \textless   step\_2\textgreater\textless rationale\_2\textgreater\textless type\_name\_2\textgreater\textless /step\_2\textgreater\\
        ...
    
        [Final answer] \\
        
        \textless step\_1\textgreater\textless type\_name\_1\textgreater\textless /step\_1\textgreater\\
        \textless step\_2\textgreater\textless type\_name\_2\textgreater\textless /step\_2\textgreater\\
        ...
    \end{mdframed}
    \captionsetup{hypcap=false}
    \captionof{figure}{The prompt template for the classifying each steps into four habits classes.}
    \label{fig:classify_prompt}
\end{minipage}

\pagebreak
\subsection{Comparison of Segmentation-Classification and Prompt-base Counting Method }
\label{app:compare_classify_method_baseline}

In this section, we showcase the behavior calculated by the prior work \cite{gandhi2025cognitive} using counting prompt and compared to our segmentation-classification method (seg-class). As seen in the results, our result always resulted in lower behavior numbers than Counting method.
\begin{table}[t]
  \centering
  \small    \caption{Comparison between Counting \citep{gandhi2025cognitive} and \textbf{seg-class (ours)} methods for \texttt{R1-Distill-Llama-8B} on \textsc{MATH-500} benchmark (English problem statements; generation prefixed with target language)}  \label{tab:math500_comparison_transposed_no_accuracy}
  \begin{tabular}{lcccccccc}
    \toprule
    \textbf{Lang} & \multicolumn{2}{c}{\textbf{Subgoal setting}} & \multicolumn{2}{c}{\textbf{Backtracking}} & \multicolumn{2}{c}{\textbf{Verification}} & \multicolumn{2}{c}{\textbf{Backward chaining}} \\
    & \textbf{Count} & \textbf{seg-class} & \textbf{Count} & \textbf{seg-class} & \textbf{Count} & \textbf{seg-class} & \textbf{Count} & \textbf{seg-class} \\
    \midrule
    \textbf{En} & 6.02 & \textbf{2.73} & 4.66 & \textbf{0.76} & 6.90 & \textbf{7.27} & 2.45 & \textbf{0.016} \\
    \textbf{Zh} & 6.83 & \textbf{3.26} & 5.89 & \textbf{0.65} & 7.76 & \textbf{8.45} & 2.49 & \textbf{0.018} \\
    \textbf{Es} & 3.67 & \textbf{1.81} & 0.76 & \textbf{0.18} & 1.61 & \textbf{0.34} & 0.48 & \textbf{0.0} \\
    \textbf{Ru} & 6.46 & \textbf{2.84} & 5.27 & \textbf{0.84} & 6.67 & \textbf{5.34} & 2.87 & \textbf{0.006} \\
    \textbf{Ja} & 5.08 & \textbf{2.97} & 1.87 & \textbf{0.60} & 8.53 & \textbf{3.58} & 0.81 & \textbf{0.004} \\
    \textbf{Ko} & 5.29 & \textbf{2.36} & 2.58 & \textbf{0.39} & 4.82 & \textbf{5.06} & 1.65 & \textbf{0.006} \\
    \textbf{Te} & 2.67 & \textbf{0.68} & 1.29 & \textbf{0.17} & 2.08 & \textbf{1.11} & 1.36 & \textbf{0.0} \\
    \textbf{Sw} & 4.62 & \textbf{1.32} & 1.51 & \textbf{0.23} & 4.07 & \textbf{1.33} & 1.58 & \textbf{0.011} \\
    \bottomrule
  \end{tabular}
\end{table}
\clearpage
\section{Limiting Tokens to Control Output Language}
\label{app:limit_language_via_masking}

We explore the idea of limiting the available allowed tokens during decoding to force LRM to output in a certain language. This solution would allows more freedom of what kind of reasoning compare to our method which seeds the initial reasoning with a opening phrase.

We first indentify tokens which uses to generate our target languages from Distill-Llama-8B. In Llama 3 tokenizers, we found 4,225 tokens are Chinese text generation, 1,410 tokens are related to Japanese text generation. The low amount of Japanese tokens may limits the capabilities of final results as LLMs can only output from only 1410 tokens. This exposed the limitations of using masking as a way to limit reasoning language.

\begin{table}[htbp]
\centering
\caption{Results on Llama-8B with Japanese prefill with ``\begin{CJK}{UTF8}{min}まず\end{CJK}'' and ``\begin{CJK}{UTF8}{gbsn}嗯\end{CJK}'', comparing Prefill Target Language and }
\begin{tabular}{lcc}
\toprule
\textbf{Target Language} & \textbf{Japanese (\%)} & \textbf{Chinese (\%)} \\
\midrule
Input=English, Prefill Target Language & 64.8 & 67.8 \\
\quad - Thinking language (en/zh/native/no) & 0.2 / 75.2 / 0.2 / 17.4 & 0.0 / 74.2 / 0.0 / 25.8 \\
\quad - Answer language (en/zh/native/no) & 4.4/69.8/4.4/17.4 & 0.8/71.4/0.8/25.8 \\
\midrule
Input=English, Masking non Target Tokens & 61.4 & 69.6 \\
\quad - Thinking language (en/zh/native/no) & 15.2/12.6/15.2/18.6 & 6.6/80.0/6.6/13.4 \\
\quad - Answer language (en/zh/native/no) & 35.0/18.8/35.0/18.6 & 8.2/77.0/8.2/13.2 \\
\midrule
Input=Target, Prefill Target Language & 32.6 & 73.6 \\
\quad - Thinking language (en/zh/native/no) & 0.2/14.8/75.0/9.8 & 0.0/73.8/73.8/26.2 \\
\quad - Answer language (en/zh/native/no) & 2.4/18.4/67.2/9.8 & 0.2/71.6/71.6/26.2 \\
\midrule
Input=Target, Masking non Target Tokens  & 42.0 & 73.6 \\
\quad - Thinking language (en/zh/native/no) & 0.6/18.4/63.6/17.0 & 0.0/92.0/92.0/8.0 \\
\quad - Answer language (en/zh/native/no) & 4.6/21.2/54.6/17.0 & 0.2/89.4/89.4/8.0 \\
\bottomrule
\end{tabular}
\label{tab:language_comparison}
\end{table}

\clearpage
\section{Prefill Phrases}
\label{app:prefill_phrase_study}

\subsection{Distribution Found From MATH-500 Baseline}
\label{app:baseline_setting_distribution}

To find the distribution of prefill tokens across different languages and models, we analyzed the output generations from multiple language models on a subset of the MATH-500 baseline dataset. For each model and language combination, we recorded the first n tokens generated (where n=4 in our analysis) and tracked their frequencies across all sampled problems.

We implemented a token tracking system that builds up sequences by concatenating successive tokens (e.g., first token, first+second tokens, etc.) and maintains frequency counts for each unique sequence at each position.
For models where we had access to the tokenizer, we performed additional analysis by converting between token IDs and human-readable text, allowing us to identify meaningful phrases rather than just token sequences. This double decoding process was particularly valuable for non-Latin script languages where token boundaries might not align with linguistic units. The resulting distributions, shown in Table~\ref{tab:frequent-phrases-math-500}.

\begin{table}[t]
\caption{Most Frequent Starting Phrases by Model and Language, (-) indicate using the most common prefill target phrase from other models.}
\label{tab:frequent-phrases-math-500}
\begin{center}
\begin{tabular}{lcccc}
\toprule
Model & Language & Most Frequent Phrase & Count & Representative Phrase (Count) \\
\midrule
R1-Distill-Llama-8B & es & Okay & 248 & Primero (224) \\
R1-Distill-Llama-8B & sw & Okay & 253 & Mama (62) \\
R1-Distill-Llama-8B & en & Okay & 451 & Okay (451) \\
R1-Distill-Llama-8B & ja & \begin{CJK}{UTF8}{gbsn}好，我现在要\end{CJK} & 196 & \begin{CJK}{UTF8}{min}まず\end{CJK} (112) \\
R1-Distill-Llama-8B & ko & \begin{CJK}{UTF8}{gbsn}首先，我需要\end{CJK} & 107 & \begin{CJK}{UTF8}{mj}먼저\end{CJK} (35) \\
R1-Distill-Llama-8B & ru & \foreignlanguage{russian}{Хорошо} & 130 & \foreignlanguage{russian}{Хорошо} (130) \\
R1-Distill-Llama-8B & zh-CN & \begin{CJK}{UTF8}{gbsn}嗯\end{CJK} & 305 & \begin{CJK}{UTF8}{gbsn}嗯\end{CJK} (305) \\
\midrule
Qwen-14B & es & Okay, & 209 & Primero (166) \\
Qwen-14B & sw & Okay, so I & 173 & Kwa (43) \\
Qwen-14B & en & Okay, & 345 & Okay (345) \\
Qwen-14B & ja & \begin{CJK}{UTF8}{gbsn}好，\end{CJK} & 204 & \begin{CJK}{UTF8}{min}まず\end{CJK} (150) \\
Qwen-14B & ko & \begin{CJK}{UTF8}{gbsn}嗯，\end{CJK} & 204 & \begin{CJK}{UTF8}{mj}먼저\end{CJK} (78) \\
Qwen-14B & ru & \foreignlanguage{russian}{Хорошо} & 278 & \foreignlanguage{russian}{Хорошо} (386) \\
Qwen-14B & zh-CN & \begin{CJK}{UTF8}{gbsn}首先，\end{CJK} & 181 & \begin{CJK}{UTF8}{gbsn}首先\end{CJK} (181) \\
\midrule
QwQ-32B & es & Okay, & 489 & Primero (3) \\
QwQ-32B & sw & Okay, & 474 & Ili kup (2) \\
QwQ-32B & en & Okay, & 477 & Okay, (477) \\
QwQ-32B & ja & Alright, & 208 & \begin{CJK}{UTF8}{min}まず\end{CJK} (123) \\
QwQ-32B & ko & \begin{CJK}{UTF8}{mj}좋아\end{CJK} & 220 & \begin{CJK}{UTF8}{mj}좋아\end{CJK} (220) \\
QwQ-32B & ru & \foreignlanguage{russian}{Хорошо} & 365 & \foreignlanguage{russian}{Хорошо} (365) \\
QwQ-32B & zh-CN & \begin{CJK}{UTF8}{gbsn}嗯，\end{CJK} & 479 & \begin{CJK}{UTF8}{gbsn}嗯，\end{CJK} (479) \\
QwQ-32B & te & Okay, & 499 & prārambhiṃcaḍāniki (-)
 \\
\midrule
Qwen3-30B-A3B & es & Okay, & 490 & Primero (5) \\
Qwen3-30B-A3B & sw & Okay, & 494 & Ili kup (1) \\
Qwen3-30B-A3B & en & Okay, & 487 & Okay, (487) \\
Qwen3-30B-A3B & ja & Okay, & 491 & \begin{CJK}{UTF8}{min}まず\end{CJK} (5) \\
Qwen3-30B-A3B & te & Okay, & 499 & prārambhiṃcaḍāniki (3) \\
Qwen3-30B-A3B & ko & Okay, & 491 & \begin{CJK}{UTF8}{mj}좋아\end{CJK} (2) \\
Qwen3-30B-A3B & ru & \foreignlanguage{russian}{Хорошо} & 490 & \foreignlanguage{russian}{Хорошо} (490) \\
Qwen3-30B-A3B & zh-CN & \begin{CJK}{UTF8}{gbsn}嗯，\end{CJK} & 487 & \begin{CJK}{UTF8}{gbsn}嗯，\end{CJK} (487) \\
\bottomrule
\end{tabular}
\end{center}
\end{table}

\pagebreak
\subsection{CulturalBench Prefill Phrase}
\label{app:prefill_phrase_cultural_bench}

The following Table \ref{tab:cultural_bench_phrase} showcase the phrases used to prefill target language in CultureBench-Hard.

\begin{table}[htbp]
    \centering
    \caption{Preferred prefill tokens used by language models across different countries, reflecting culturally-specific conversational cues.}
    \label{tab:cultural_bench_phrase}
    \begin{tabular}{@{}ll@{}}
        \toprule
        \textbf{Country} & \textbf{Prefill token} \\
        \midrule
        Argentina        & Vale                                               \\
        Australia        & Okay                                               \\
        Brazil           & Tudo bem                                           \\
        Canada           & Okay                                               \\
        Chile            & Vale                                               \\
        China            & \begin{CJK}{UTF8}{gbsn}嗯\end{CJK}                 \\
        Czech Republic   & Dobře                                              \\
        France           & D'accord                                           \\
        Germany          & In Ordnung                                         \\
        Hong Kong        & \begin{CJK}{UTF8}{gbsn}嗯\end{CJK}                 \\
        Indonesia        & Baiklah                                            \\
        Italy            & Va bene                                            \\
        Japan            & \begin{CJK}{UTF8}{min}まず\end{CJK}                \\
        Malaysia         & Baiklah                                            \\
        Mexico           & Órale                                              \\
        Netherlands      & Oké                                                \\
        New Zealand      & Okay                                               \\
        Nigeria          & Okay                                               \\
        Peru             & Ya                                                 \\
        Philippines      & Sige                                               \\
        Poland           & Dobrze                                             \\
        Romania          & Bine                                               \\
        Russia           & \foreignlanguage{russian}{Хорошо}                  \\
        Singapore        & Okay                                               \\
        South Africa     & Okay                                               \\
        South Korea      & 먼저                                                \\
        Spain            & Vale                                               \\
        Taiwan           & \begin{CJK}{UTF8}{gbsn}嗯\end{CJK}                 \\
        Turkey           & Tamam                                              \\
        Ukraine          & \foreignlanguage{russian}{Добре}                   \\
        United Kingdom   & Alright                                            \\
        United States    & Okay                                               \\
        Zimbabwe         & Okay                                               \\
        \bottomrule
    \end{tabular}
\end{table}
\clearpage
\section{Additional MATH-500 Details}

Table \ref{tab:math-500-prefill-comparison_with_baseline} shows the scores of all models with the inclusion of baseline score.

\begin{table}
\centering
\caption{Comparison of English vs. Native prefill strategies on MATH-500 across languages order by speakers population.}
\label{tab:math-500-prefill-comparison_with_baseline}
\small
\begin{tabular}{lccccccc}
\toprule
Strategy & Chinese & Spanish & Russian & Swahili & Japanese & Telugu & Korean \\
\midrule
\multicolumn{8}{c}{\textbf{DeepSeek-R1-Distill-Llama-8B}} \\
English Prefill & 78.8\% & 80.2\% & 78.4\% & 37.0\% & 74.6\% & 42.2\% & 69.8\% \\
Native Prefill & 73.6\% & 45.6\% & 59.4\% & 3.8\% & 32.6\% & 16.8\% & 41.6\% \\
Baseline & 75.8\% & 70.6\% & 69.8\% & 27.3\% & 61.2\% & 41.6\% & 64.6\% \\
Difference (EN - Native) & \textbf{+5.2\%} & \textbf{+34.6\%} & \textbf{+19.0\%} & \textbf{+33.2\%} & \textbf{+42.0\%} & \textbf{+25.4\%} & \textbf{+28.2\%} \\
\midrule
\multicolumn{8}{c}{\textbf{DeepSeek-R1-Distill-Qwen-14B}} \\
English Prefill & 88.4\% & 88.6\% & 86.6\% & 52.4\% & 85.2\% & 66.2\% & 84.4\% \\
Native Prefill & 89.8\% & 66.4\% & 86.4\% & 14.6\% & 63.6\% & 34.4\% & 83.8\% \\
Baseline & 82.6\% & 88.0\% & 84.6\% & 39.8\% & 80.0\% & 64.6\% & 83.4\% \\
Difference (EN - Native) & -1.4\% & \textbf{+22.2\%} & \textbf{+0.2\%} & \textbf{+37.8\%} & \textbf{+21.6\%} & \textbf{+31.8\%} & \textbf{+0.6\%} \\
\midrule
\multicolumn{8}{c}{\textbf{QwQ-32B}} \\
English Prefill & 92.4\% & 92.2\% & 91.2\% & 67.8\% & 90.2\% & 84.4\% & 90.6\% \\
Native Prefill & 90.6\% & 93.2\% & 90.6\% & 55.6\% & 87.4\% & 65.2\% & 88.2\% \\
Baseline & 90.8\% & 93.2\% & 90.8\% & 68.2\% & 89.4\% & 85.0\% & 89.0\% \\
Difference (EN - Native) & \textbf{+1.8\%} & -1.0\% & \textbf{+0.6\%} & \textbf{+12.2\%} & \textbf{+2.8\%} & \textbf{+19.2\%} & \textbf{+2.4\%} \\
\midrule
\multicolumn{8}{c}{\textbf{Qwen3-30B-A3B}} \\
English Prefill & 91.4\% & 91.0\% & 90.6\% & 72.4\% & 89.4\% & 87.0\% & 89.8\% \\
Native Prefill & 89.4\% & 83.8\% & 90.0\% & 29.6\% & 81.8\% & 68.4\% & 88.0\% \\
Baseline & 89.4\% & 91.2\% & 90.4\% & 72.8\% & 90.0\% & 87.7\% & 89.2\% \\
Difference (EN - Native) & \textbf{+2.0\%} & \textbf{+7.2\%} & \textbf{+0.6\%} & \textbf{+42.8\%} & \textbf{+7.6\%} & \textbf{+18.6\%} & \textbf{+1.8\%} \\
\midrule
\multicolumn{8}{c}{\textbf{Average across all models}} \\
English Prefill & 87.7\% & 88.0\% & 86.7\% & 57.4\% & 84.9\% & 70.0\% & 83.7\% \\
Native Prefill & 85.9\% & 72.2\% & 81.6\% & 25.9\% & 66.3\% & 46.2\% & 75.4\% \\
Baseline & 84.6\% & 85.8\% & 83.9\% & 52.0\% & 80.2\% & 69.7\% & 81.5\% \\
Difference (EN - Native) & \textbf{+1.9\%} & \textbf{+15.8\%} & \textbf{+5.1\%} & \textbf{+31.5\%} & \textbf{+18.5\%} & \textbf{+23.7\%} & \textbf{+8.3\%} \\
\bottomrule
\end{tabular}
\end{table}
\clearpage
\section{MMMLU Results}
\label{app:mmmlu_additional}

\subsection{MMMLU full models breakdown}

Table \ref{tab:mmlu-reasoning-language-comparison_full} shows the full results for four models. We observe a significant jump in QwQ-32B where switching Swahili MMLU from English reasoning to Swahili reasoning drops by over 36\%.

\begin{table}
\centering
\caption{Comparison of MMLU performance when reasoning in native language vs. English}
\label{tab:mmlu-reasoning-language-comparison_full}
\small
\begin{tabular}{lcccccc}
\toprule
Strategy & English & Chinese & Spanish & Swahili & Japanese & Korean \\
\midrule
\multicolumn{7}{c}{\textbf{R1-Distill-Llama-8B}} \\
Prefill English & -- & 69.8\% & 71.4\% & 29.8\% & 65.3\% & 61.5\% \\
Prefill target Language & 67.7\% & 63.4\% & 53.8\% & 18.6\% & 46.2\% & 46.8\% \\
Difference (EN - Native) & -- & \textbf{+6.4\%} & \textbf{+17.6\%} & \textbf{+11.2\%} & \textbf{+19.1\%} & \textbf{+14.6\%} \\
\midrule
\multicolumn{7}{c}{\textbf{Qwen-14B}} \\
Prefill English & -- & 84.7\% & 85.7\% & 44.5\% & 83.3\% & 81.1\% \\
Prefill target Language & 87.3\% & 83.3\% & 85.8\% & 36.4\% & 77.3\% & 73.4\% \\
Difference (EN - Native) & -- & \textbf{+1.4\%} & -0.1\% & \textbf{+8.1\%} & \textbf{+5.9\%} & \textbf{+7.7\%} \\
\midrule
\multicolumn{7}{c}{\textbf{QwQ-32B}} \\
Prefill English & -- & 88.7\% & 89.1\% & 59.8\% & 87.8\% & 85.8\% \\
Prefill target Language & 91.4\% & 88.5\% & 89.2\% & 23.8\% & 88.3\% & 83.6\% \\
Difference (EN - Native) & -- & \textbf{+0.3\%} & -0.1\% & \textbf{+36.0\%} & -0.5\% & \textbf{+2.2\%} \\
\midrule
\multicolumn{7}{c}{\textbf{Qwen3-30B-A3B}} \\
Prefill English & -- & 88.9\% & 89.0\% & 61.1\% & 86.8\% & 82.1\% \\
Prefill target Language & 85.4\% & 85.6\% & 86.0\% & 62.2\% & 84.2\% & 81.0\% \\
Difference (EN - Native) & -- & \textbf{+3.3\%} & \textbf{+3.0\%} & -1.0\% & \textbf{+2.7\%} & \textbf{+1.1\%} \\
\midrule
\multicolumn{7}{c}{\textbf{Average across all models}} \\
Prefill English & -- & 83.1\% & 83.8\% & 48.8\% & 80.8\% & 77.6\% \\
Prefill target Language & 83.0\% & 80.2\% & 78.7\% & 35.3\% & 74.0\% & 71.2\% \\
Difference (EN - Native) & -- & \textbf{+2.9\%} & \textbf{+5.1\%} & \textbf{+13.6\%} & \textbf{+6.8\%} & \textbf{+6.4\%} \\
\bottomrule
\end{tabular}
\end{table}

\subsection{Scores in subset versus full set}
Table \ref{tab:mmmlu-comparison-full-subset} showcases the accuracy between the 32 subjects and the full 56 subjects score. All settings consistently score higher than the full set; however, the correlation score between different settings is 0.9953 with a p-value lower than 0.0001. This means the subsets we have chosen are representative enough of the full MMMLU test set.

\begin{table}[t]
\caption{Comparison of MMMLU partial (subset) and full accuracy scores across different models and language configurations.}
\label{tab:mmmlu-comparison-full-subset}
\centering
\begin{tabular}{lccccc}
\toprule
\textbf{Model} & \textbf{Input} & \textbf{Reasoning} & \textbf{Partial Acc.} & \textbf{Full Acc.} & \textbf{Diff.} \\
\midrule
DeepSeek-R1-Distill-Qwen-14B & en & en & 88.02\% & 85.61\% & +2.41\% \\
QwQ-32B & en & es & 91.04\% & 88.52\% & +2.52\% \\
DeepSeek-R1-Distill-Qwen-14B & en & zh-CN & 86.63\% & 84.09\% & +2.54\% \\
DeepSeek-R1-Distill-Qwen-14B & en & ko & 85.40\% & 82.52\% & +2.88\% \\
DeepSeek-R1-Distill-Qwen-14B & es & en & 85.69\% & 82.68\% & +3.01\% \\
DeepSeek-R1-Distill-Qwen-14B & en & es & 84.63\% & 82.12\% & +2.51\% \\
DeepSeek-R1-Distill-Qwen-14B & en & ja & 83.74\% & 81.29\% & +2.45\% \\
DeepSeek-R1-Distill-Qwen-14B & zh-CN & zh-CN & 83.64\% & 80.33\% & +3.31\% \\
DeepSeek-R1-Distill-Qwen-14B & ja & en & 83.26\% & 79.97\% & +3.29\% \\
DeepSeek-R1-Distill-Qwen-14B & ko & en & 81.08\% & 78.02\% & +3.06\% \\
\bottomrule
\end{tabular}
\end{table}
\clearpage
\section{CulturalBench Results}
\label{app:culturalbench_details}

In CulturalBench, we maintained the original English questions while only varying the reasoning language. This approach preserves the precise wording of questions, as translation could potentially compromise the cultural nuances embedded in specific English terminology unique to each culture.

\begin{figure}[t]
    \centering
    \includegraphics[width=\textwidth]{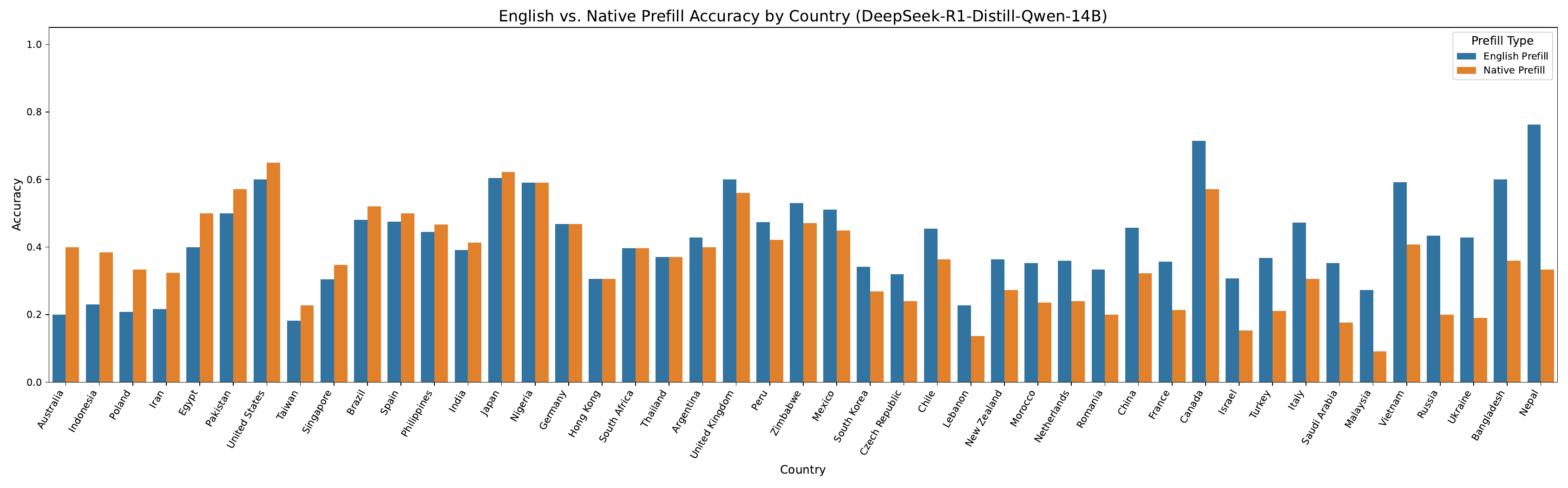}
    \caption{Sorted by positive improvements from using native language reasoning compare to english reasoning in Deepseek-Distill-Qwen-14B}
    \label{fig:culture_improvement_breakdown_by_country_qwen14}
\end{figure}

\begin{figure}[t]
    \centering
    \includegraphics[width=\textwidth]{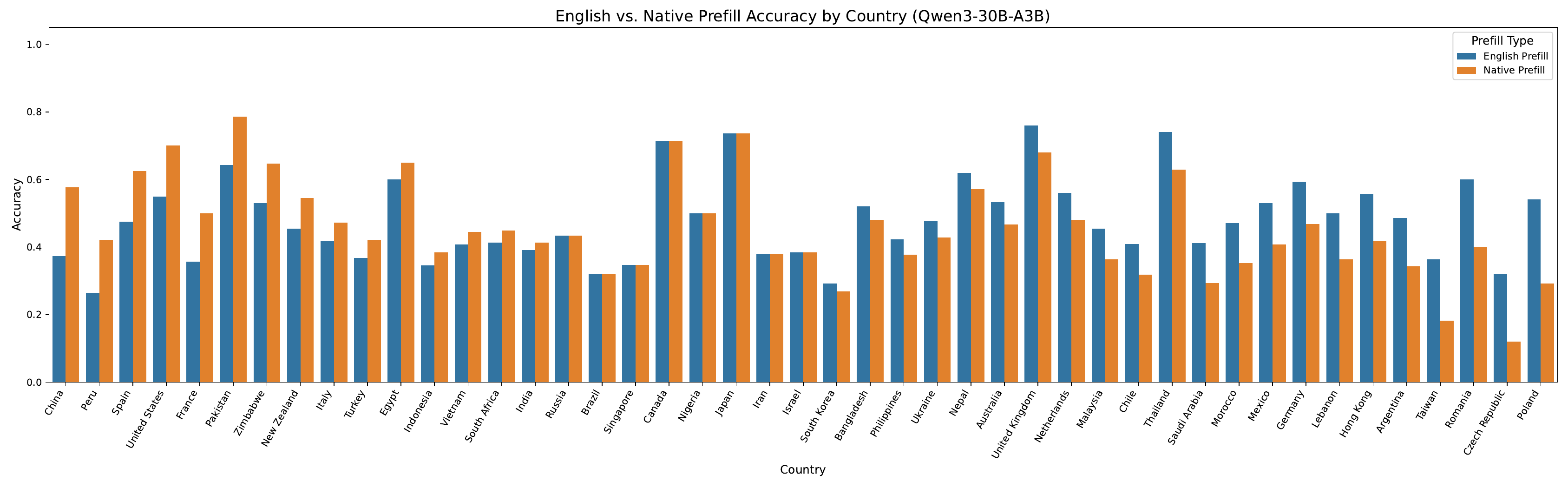}
    \caption{Sorted by positive improvements from using native language reasoning compare to english reasoning in Qwen3-30B-A3B}
    \label{fig:culture_improvement_breakdown_by_country_qwen3}
\end{figure}

\begin{figure}[t]
    \centering
    \includegraphics[width=\textwidth]{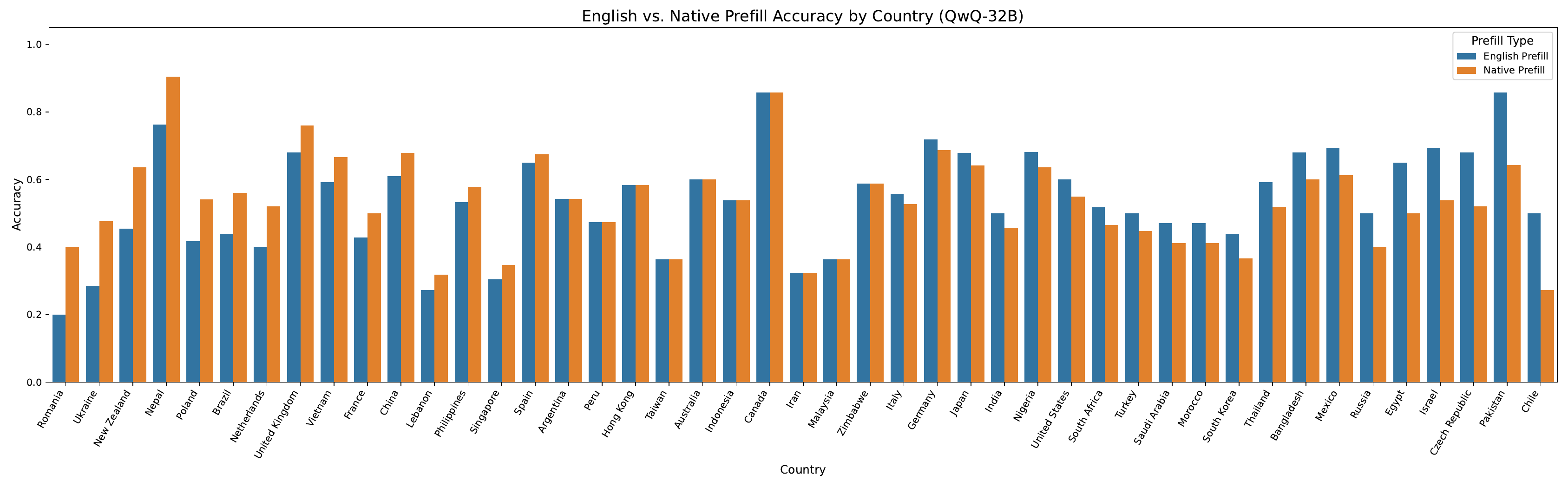}
    \caption{Sorted by positive improvements from using native language reasoning compare to english reasoning in QwQ-32B}
    \label{fig:culture_improvement_breakdown_by_country_qwq}
\end{figure}

Figures \ref{fig:culture_improvement_breakdown_by_country_qwen14}, \ref{fig:culture_improvement_breakdown_by_country_qwen3}, and \ref{fig:culture_improvement_breakdown_by_country_qwq} illustrate the performance difference between using English prefills versus prefills in the predominant language of each respective country. 

\clearpage
\section{Study of Impact of Prefill Tokens in Pretrained Model}
\label{app:base_model_aime}
To investigate why models might gravitate towards English and Chinese for reasoning, we conducted an experiment using a small mathematics problem set, AIME-2024. Using the prompt template from Deepseek-R1-zero \citep{guo2025deepseek}, we prompted the Qwen3-30B-A3B base model (without post-training) in a zero-shot pass@8 setting. To encourage reasoning in languages other than English, we prepended an initial phrase in the target language to the prompt, guiding the model to complete its reasoning in that language. The results, presented in Table \ref{tab:aime-base-zero-prompt-scores}, show that English-led reasoning significantly outperforms other languages for this base model.

\begin{table}[t]
\caption{AIME-24 pass@8 from Qwen3-30B-A3B \textbf{base} model with different initial phrase for text completion.}
\label{tab:aime-base-zero-prompt-scores}
\centering
\begin{tabular}{lcccccc}
\toprule
 & \multicolumn{6}{c}{\textbf{Language}} \\
\cmidrule(lr){2-7}
 & en & zh-CN & ja & ru & ko & sw \\
\midrule
Phrase & Okay & \begin{CJK}{UTF8}{gbsn}嗯\end{CJK} & \begin{CJK}{UTF8}{min}まず\end{CJK} & \foreignlanguage{russian}{Хорошо} & 먼저 & Kwa kuzingatia \\
pass@8 & \textbf{0.267} & 0.190 & 0.172 & 0.133 & 0.133 & 0.200 \\
\bottomrule
\end{tabular}
\end{table}

Based on these findings, we hypothesize that during the RL training phase, models tend to exploit the language that allows the most effective CoT generation to maximize the final task score. Since the choice of reasoning language is typically not an explicit part of the reward function, leveraging the language in which the underlying base model performs best (as suggested by Table \ref{tab:aime-base-zero-prompt-scores} for English) becomes an optimal strategy for achieving higher rewards.
\clearpage
\section{Brittleness of language guidance in LRM compared to typical CoT found in LLMs}

\begin{table}[t]
\caption{Comparison prefilling reasoning chain with native language or english in reasoning on, while prefilling the response in reasoning off, meaning the model does not undergo long CoT process before output response.}
\label{tab:reasoning_vs_cot_comparison}
\centering
\begin{tabular}{lcccccc}
\toprule
MATH-500 & \multicolumn{6}{c}{Language} \\
\cmidrule(r){2-7}
Model Configuration & Chinese & Japanese & Korean & Spanish & Russian & Telugu \\
\midrule
\multicolumn{7}{l}{\textit{Qwen3 30 A3B (reasoning off)}} \\
Prefill English (To evaluate) & 84.8\% & 81.6\% & 80.0\% & 80.2\% & 82.2\% & 81.2\% \\
Prefill Input Language & 88.4\% & 80.8\% & 82.2\% & 81.2\% & 79.4\% & 68.3\% \\
\rowcolor{lightblue}
Difference (English - Input) & -3.6\% & 0.8\% & -2.2\% & 1.0\% & 2.8\% & 12.9\% \\
\midrule
\multicolumn{7}{l}{\textit{Qwen3 30 A3B (reasoning on)}} \\
Prefill English (Okay) & 91.4\% & 89.4\% & 89.8\% & 91.0\% & 90.6\% & 87.0\% \\
Prefill Input Language & 89.4\% & 81.8\% & 88.0\% & 83.8\% & 90.0\% & 68.4\% \\
\rowcolor{lightblue}
Difference (English - Input) & 2\% & 7.6\% & 1.8\% & 7.2\% & 0.6\% & 18.6\% \\
\bottomrule
\end{tabular}
\end{table}

Since Qwen3-30B-A3B allows us to trigger reasoning mode on and off, we first compare the sensitivity between reasoning and normal CoT prompts. Specially we compare the results between prefilling the phrase in reasoning versus preflling in the response in CoT response with reasoning mode off. Table \ref{tab:reasoning_vs_cot_comparison} shows that the penalty of changing reasoning language is far more worse than changing in typical chain of thought from LLMs. 

\clearpage
\section{Dataset Details}
\label{app:dataset_details_license}
Table \ref{tab:dataset_licenses} contains each of the benchmarks and their licenses.

\begin{table}[t]
\centering
\small
\begin{tabular}{lcl}
\toprule
\textbf{Dataset} & \textbf{Test Split Size} & \textbf{License} \\
\midrule
MMMLU$^1$ & N = 14,042 (per language) & MIT License \\
CulturalBench-Hard$^2$ & N = 4,709 & CC-BY-4.0 \\
LMSYS-toxic$^3$ & N = 2,000  (per language) & LMSYS-Chat-1M Dataset License Agreement \\
MATH-500$^4$ & N = 500 (per language) & MIT License \\
\bottomrule
\end{tabular}
\caption{AI Dataset Information with Test Split Sizes}
\label{tab:dataset_licenses}
\end{table}

\textbf{Languages:}
\begin{compactitem}
\item MMMLU: English, Spanish, Japanese, Korean, Swahili, Chinese
\item CulturalBench-Hard: 30 countries
\item LMSYS-toxic: English, Japanese, Spanish, Korean, Swahili, Telugu, Russian, Chinese
\item MATH-500: English, Japanese, Korean, Spanish, Swahili, Telugu, Russian, Chinese
\end{compactitem}
\textbf{HuggingFace Link:}\
$^1$ MMMLU: \url{https://huggingface.co/datasets/openai/MMMLU}\\
$^2$ CulturalBench: \url{https://huggingface.co/datasets/kellycyy/CulturalBench}\\
$^3$ LMsys-Chat-1M: \url{https://huggingface.co/datasets/lmsys/lmsys-chat-1m}\\
$^4$ MATH-500: \url{https://huggingface.co/datasets/HuggingFaceH4/MATH-500}
\clearpage
\section{Behavior Results Detail for MATH-500}
\label{app:full_behavior_results}

This section details the behavioral results observed for the MATH-500 dataset, specifically examining the correlation between language and various reasoning behaviors. The analysis, as presented in Tables \ref{tab:prefill_lang_corr} and \ref{tab:input_lang_corr}, investigates how different languages, when used either as prefill tokens to guide the model's internal ``thought'' process or as the input language of the problems themselves, influence reasoning strategies such as backtracking, backward chaining, subgoal setting, and verification. Notably, Chinese (zh-CN) prefill tokens show a strong positive correlation with subgoal setting ($r=0.50, p<0.001$) and verification ($r=0.41, p<0.001$). Conversely, English prefill is significantly positively correlated with backward chaining ($r=0.34, p<0.01$), while Swahili shows a significant negative correlation with subgoal setting ($r=-0.35, p<0.01$) when used as a prefill language. When considering input languages, Chinese again demonstrates a significant positive correlation with subgoal setting ($r=0.42, p<0.001$) and verification ($r=0.33, p<0.01$). These findings suggest that linguistic context, whether from prefill or input, can systematically influence the reasoning patterns employed by the models when tackling mathematical problems.

\begin{table}[t]
\caption{Correlation between \textbf{prefill target languages} and reasoning behaviors}
\label{tab:prefill_lang_corr}
\centering
\begin{tabular}{lcccc}
\toprule
Language & Backtrack & Backward & Subgoal Setting & Verification \\
\midrule
English & -0.07 & $0.34^{**}$ & 0.22 & 0.07 \\
Spanish & 0.08 & -0.16 & $-0.27^{*}$ & $-0.26^{*}$ \\
Japanese & 0.16 & -0.19 & -0.19 & $-0.29^{*}$ \\
Korean & -0.03 & 0.12 & 0.02 & 0.05 \\
Russian & -0.06 & 0.02 & 0.09 & 0.05 \\
Swahili & -0.22 & $-0.25^{*}$ & $-0.35^{**}$ & 0.16 \\
Telugu & -0.01 & -0.18 & -0.12 & -0.20 \\
zh-CN & $0.23^{*}$ & 0.02 &$ 0.50^{***}$ & $0.41^{***}$ \\
\bottomrule
\multicolumn{5}{l}{$^{*}p<0.05$, $^{**}p<0.01$, $^{***}p<0.001$}
\end{tabular}
\end{table}

\begin{table}[t]
\caption{Correlation between \textbf{input target languages} and reasoning behaviors}
\label{tab:input_lang_corr}
\centering
\begin{tabular}{lcccc}
\toprule
Language & Backtrack & Backward & Subgoal Setting & Verification \\
\midrule
English & -0.07 & 0.08 & 0.00 & 0.08 \\
Spanish & -0.07 & -0.08 & -0.19 & -0.22 \\
Japanese & 0.14 & 0.01 & -0.14 & -0.19 \\
Korean & -0.06 & 0.18 & 0.04 & 0.03 \\
Russian & -0.10 & 0.00 & 0.09 & 0.07 \\
Swahili & -0.13 & -0.10 & -0.19 & 0.09 \\
Telugu & 0.08 & -0.09 & 0.02 & -0.14 \\
zh-CN & 0.19 & 0.04 & $0.42^{***}$ & $0.33^{**}$ \\
\bottomrule
\multicolumn{5}{l}{$^{*}p<0.05$, $^{**}p<0.01$, $^{***}p<0.001$}
\end{tabular}
\end{table}
\clearpage
\pagebreak

\end{document}